\documentclass{article}

\usepackage[preprint]{neurips_2026}
\usepackage[table]{xcolor}
\usepackage[utf8]{inputenc} 
\usepackage[T1]{fontenc}    
\usepackage{hyperref}       
\usepackage{url}            
\usepackage{booktabs}       
\usepackage{amsfonts}       
\usepackage{amsmath}
\usepackage{nicefrac}       
\usepackage{microtype}      
\usepackage{xcolor}         
\usepackage{enumitem}
\usepackage{tikz}
\usepackage{graphicx}
\usepackage{subcaption}
\usepackage{wrapfig}
\usepackage{multirow}
\usepackage{alltt}
\usepackage{tcolorbox}
\usepackage{algorithm}
\usepackage{algpseudocode}
\usepackage{tablefootnote}
\title{PACE: Two-Timescale Self-Evolution for Small Language Model Agents}

\author{%
  Chen Ling\thanks{Corresponding Author}, Pei Chen, Albert Guan, Jiaming Qu, Shayan Ali Akbar,\\ \textbf{Madhu Gopinathan}, \textbf{Erwin Cornejo} \\
  Amazon\\
  \texttt{emorycl@amazon.com} \\
}

\begin{document}

\maketitle

\vspace{-3mm}
\begin{abstract}
Deploying language-model agents in production often requires substantial compute and human effort to tune prompts, parsers, validators, and other components of the agent pipeline. Self-evolution offers a promising alternative, but most existing frameworks assume access to frontier models that can reliably diagnose failures, propose revisions, and judge their own updates. We study whether frozen small language models (SLMs) can serve as effective self-evolving agents under resource constraints. We propose PACE (Prompt And Control Logic Evolution), a two-timescale framework that coordinates low-risk prompt refinement with higher-risk control-logic updates. PACE evolves prompts under fixed control logic until prompt-level gains saturate, then considers constrained control-logic updates that are accepted through held-out validation. Across three frozen SLM backbones ranging from 4B to 14B parameters and four controlled benchmarks, PACE achieves the best performance on all 12 backbone--benchmark combinations, improving over vanilla SLM agents by up to $+9.2\%$ relative improvement and over the stronger single-mode evolution baseline by up to $+5.4\%$ relative improvement. A $\tau$-bench case study further shows that PACE improves multi-turn tool-use success over vanilla and prompt-only evolution. These results suggest that reliable SLM agent self-evolution is possible without updating model weights or relying on frontier-model teachers, and that the key benefit is not any single final solver pattern but autonomous, validated discovery of task-appropriate inference strategies.
\end{abstract}

\vspace{-2mm}
\section{Introduction}
Language-model-based agents \citep{wang2024survey} have become a common abstraction for solving complex tasks through reasoning, tool use, verification, and iterative refinement. However, deploying such agents in production often requires substantial compute and repeated human intervention to tune prompts, parsers, validators, and other components of the agent pipeline. Recent work on agent self-evolution \citep{agrawal2025gepa,opsahl2024optimizing,zhang2025agentic} offers a promising alternative: agents can improve their own behavior by using execution feedback to revise \textit{prompts/task contexts}, or modify \textit{control logic} without changing the underlying model parameters. Most existing approaches, however, assume access to strong frontier models that can reliably diagnose failures, propose high-quality revisions, and judge whether those revisions should be accepted. These assumptions become fragile when the agent is powered by a small language model (SLM).

This paper studies self-evolution for frozen SLM agents. We focus on models with at most 14B parameters, where the model weights remain fixed throughout the evolution process. This setting is important for practical deployments in which local serving, latency, privacy, or cost constraints make frontier-model APIs or large-scale fine-tuning undesirable. It is also technically challenging. SLMs are more sensitive to prompt complexity and often reach diminishing returns after a small number of prompt revisions. At the same time, allowing an SLM to freely rewrite its own executable control logic\footnote{Non-parameter code or configuration that governs an agent's inference procedure, including output parsing, validation, retry/repair policies, routing or fallback rules, tool-call handling, and decoding settings. It \textit{excludes} updates to model weights.} can be unstable: proposed edits may be syntactically valid but semantically incorrect, causing silent regressions in parsing, validation, retry behavior, or inference-time decision rules.

\begin{figure}[t]
  \centering
  \includegraphics[width=0.85\linewidth]{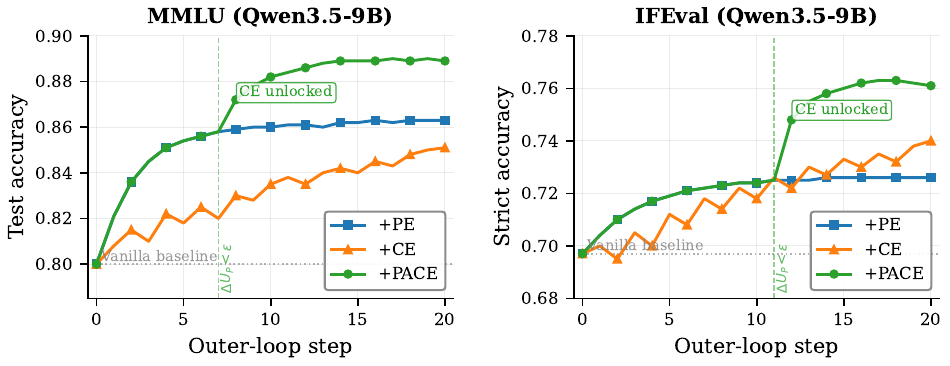}
  \vspace{-3mm}
  \caption{PACE evolution dynamics on Qwen3.5-9B. Prompt-only evolution (+PE) improves early but quickly saturates, while control-logic evolution (+CE) is noisy when structural updates are applied from the start. PACE first exploits stable prompt refinement, then introduces a validated control-logic update after prompt gains plateau, producing a performance jump and final better accuracy.}
  \label{fig:trajectory}
  \vspace{-5mm}
\end{figure}


A key observation motivating our work is that prompt updates and control-logic updates play different roles in agent improvement. As shown in Figure~\ref{fig:trajectory}, prompt-only evolution improves performance quickly but often saturates once the remaining failures stem from structural bottlenecks, such as brittle output extraction, missing validation, weak repair logic, or ineffective sampling policies. In contrast, enabling control-logic updates from the beginning is substantially less stable, since SLM-proposed structural edits can introduce regressions before the prompt has been sufficiently optimized. These results suggest that reliable SLM self-evolution should not treat prompt refinement and control-logic modification as interchangeable actions: it should first exploit low-risk prompt updates, then invoke higher-risk control-logic updates only after prompt-level gains plateau, with an explicit mechanism for validating proposed changes.

We propose \textbf{PACE} (\textbf{P}rompt \textbf{A}nd \textbf{C}ontrol Logic \textbf{E}volution), a two-timescale agentic framework for self-evolving frozen SLM. PACE operationalizes self-evolution through a controller that can invoke multiple adaptation tools, including prompt evolution, failure analysis, control logic proposal, control logic validation, etc. Prompt evolution is treated as a frequently callable, low-risk tool under a fixed agent structure, while control-logic evolution is treated as a higher-risk adaptation action that is considered if prompt-level gains saturate. To make such updates reliable, PACE separates \textit{proposal} from \textit{acceptance}: the SLM may propose constrained changes to safe-to-edit components of the agent pipeline, but a candidate structure is committed only if it improves over the current agent in a held-out validation while satisfying the resource budget. PACE is not merely prompt optimization plus code editing; the contribution is instead to show that a frozen SLM can autonomously discover when such strategies are useful, propose task-appropriate solver modifications, and commit them only through empirical validation, without human specification of the strategy–task mapping.
\begin{itemize}[nosep,leftmargin=*]
    \item We introduce PACE, a two-timescale agentic self-evolution framework for frozen SLM agents, where prompt evolution is a frequently callable low-risk adaptation tool and control-logic evolution is a higher-risk adaptation action invoked only after prompt-level gains saturate.
    \item We propose a prompt-saturation-based credit-assignment mechanism that determines when adaptation should leave prompt space and enter structural search, reducing premature code-level edits while avoiding inefficient prompt refinement after marginal gains vanish.
    \item We introduce validation-based structural evolution, where SLM-generated control-logic edits are treated as proposals and are committed only if they improve over the current agent under the held-out validation and satisfy the resource budget.
    \item We empirically validate PACE across four benchmarks and three frozen SLM backbones ranging from 4B to 14B parameters. PACE achieves the best performance on all 12 backbone--benchmark combinations, improving over the vanilla SLM baseline by relatively $9.2\%$ accuracy and over the stronger single-mode evolution baseline by relatively $5.4\%$ accuracy. Additionally, on $\tau$-bench, PACE improves multi-turn tool-use success over vanilla and prompt-only evolution.
    \end{itemize}

\section{Related Work}
Agent self-evolution \citep{tao2024survey,fang2025comprehensive} has emerged as a vital direction for improving the capabilities of language-model agents. Existing approaches can be broadly categorized according to the portion of the agent definition they are allowed to modify: 1) \textit{prompt-space evolution methods}, which operate purely in the textual domain, and 2) \textit{self-referential control logic evolution methods}, which permit modifications to executable control logic.

\textbf{Prompt-Space Agent Evolution.}
Prompt-space evolution improves agent behavior by refining textual artifacts such as system prompts, task instructions, tool descriptions, and output constraints while keeping control logic fixed. These methods use execution feedback, including incorrect outputs, reasoning traces, and tool-use errors, to guide prompt updates \citep{shinn2023reflexion, wang2023selfconsistency}. Since prompt updates remain in natural language, they are stable, sample-efficient, and easy to deploy, especially for frozen or small language models. Recent work \citep{zhang2025agentic} further treats prompt evolution as structured search, using reflection, specialization, and Pareto-aware selection to balance performance and cost \citep{agrawal2025gepa}. However, prompt-space methods cannot directly repair structural bottlenecks such as brittle parsing, missing validation, or weak retry logic, and therefore often saturate once such failures dominate.

\textbf{Control Logic Agent Evolution.} In contrast, control logic evolution allows agents to inspect and modify their own executable logic \citep{wang2024executable}, including control flow, validation routines, and inference-time configurations. These approaches are often motivated by recursive self-improvement, in which both the agent’s policy and its update mechanism evolve jointly through runtime introspection and code modification \citep{schmidhuber2003godel, yin2025godel,zhou2025multi}.

PACE differs from both lines of work in how it coordinates the two adaptation modes under frozen SLM constraints. Prompt-optimization methods generally assume a fixed execution structure, and therefore cannot directly repair structural bottlenecks such as brittle parsing or missing validation. Self-referential code-evolution methods, in contrast, often allow structural changes throughout the search process, which can be unstable when the proposing model is small \citep{lin2025se,shao2025your}. PACE addresses this gap by making the transition between prompt evolution and structural evolution explicit: structural search is delayed until prompt gains saturate, and accepted only through a validation gate. More broadly, PACE treats prompt and control-logic updates not as independent optimization targets, but as adaptation actions with different risks that must be scheduled and validated under the limited proposal quality of frozen SLMs. Thus, the novelty of PACE lies not in allowing both prompt and control-logic changes, but in assigning them to different timescales and validating the higher-risk structural updates before committing them.

\section{PACE: Two-Timescale Self-Evolution for Frozen SLM Agents}
We consider self-evolution for agents powered by frozen, resource-constrained language models. We first define the agent optimization objective, then introduce PACE as a two-timescale framework for coordinating prompt refinement and constrained control-logic updates.

\subsection{Problem Definition and Objective}

Let $\mathcal{T}$ denote a distribution over tasks, and let $M_\theta$ be a pretrained language model with fixed parameters $\theta$. Throughout evolution, $\theta$ remains frozen: the model is not fine-tuned, distilled, or otherwise updated. Adaptation is restricted to the agent definition around the model.

We define an agent as $A=(P,C)$, where $P$ denotes textual artifacts such as system prompts, task instructions, and formatting constraints, and $C$ denotes executable control logic such as parsing routines, validation modules, fallback strategies, and inference-time configurations. Given a task $\tau \sim \mathcal{T}$, the agent produces $y=A(\tau;M_\theta,P,C)$ and is evaluated by a task-specific utility $U(\tau,y)$. Executing the agent incurs a cost $\mathrm{Cost}(A)$, such as latency, token usage, model calls, or API calls. The goal of self-evolution is to iteratively improve $(P,C)$ under a fixed resource budget $B$:
\begin{equation}
\max_{P, C}
\;\;
\mathbb{E}_{\tau \sim \mathcal{T}}
\left[
U(\tau, A(\tau; M_\theta, P, C))
\right]
\quad
\text{s.t. }
\mathrm{Cost}(A) \leq B .
\label{eq:global-objective}
\end{equation}
Because $M_\theta$ remains fixed, all performance gains must arise from changes to $(P, C)$.

\begin{figure}[t]
    \centering
    \includegraphics[width=0.75\linewidth]{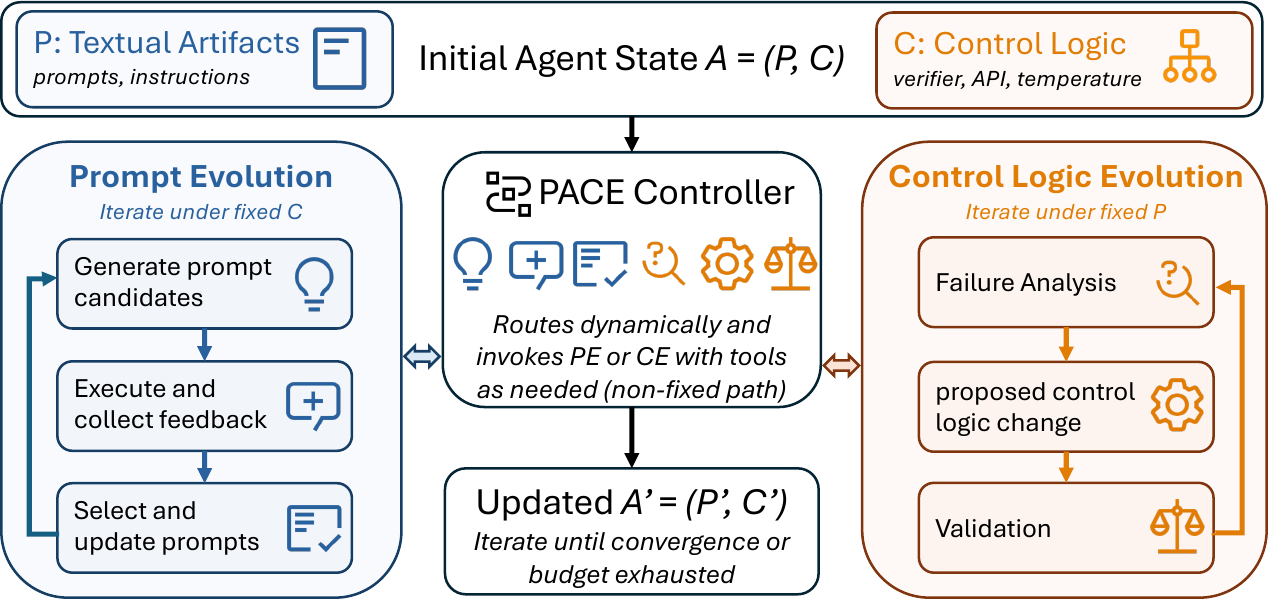}
    \caption{Overview of PACE. An agentic controller invokes prompt evolution until gains saturate, then proposes bounded control-logic updates and commits them only after held-out validation.}
    \label{fig:PACE_framework}
    \vspace{-3mm}
\end{figure}

\subsection{PACE: A Two-Timescale Agentic Adaptation Framework}
\label{sec:PACE_framework}
Directly optimizing $(P,C)$ in Eq.~\eqref{eq:global-objective} is difficult under frozen SLM constraints because prompt edits and control-logic edits fail in different ways. Prompt evolution is inexpensive and stable, but it can saturate once failures stem from structural bottlenecks such as brittle parsing, missing validation, weak retry logic, or ineffective sampling. Control-logic evolution can address these failures by changing how model outputs are sampled, parsed, checked, repaired, or re-executed, but such edits are higher-risk: SLM-proposed changes may be non-executable or semantically incorrect.

PACE therefore treats self-evolution as a two-timescale agentic adaptation process rather than an exact joint optimizer over $(P,C)$. As summarized in Figure~\ref{fig:PACE_framework}, an agentic controller invokes prompt evolution as the default low-risk tool, uses failure analysis once prompt gains saturate, proposes bounded control-logic updates, and commits a candidate solver only if it improves held-out validation performance under the resource budget. After each accepted update, PACE re-invokes prompt evolution under the new control logic. This cycle coordinates low-risk prompt refinement with higher-risk structural adaptation while reducing premature or harmful control-logic changes.


\textbf{Prompt Evolution (PE).}
For a fixed control logic $C$, PACE uses prompt evolution as the primary low-risk mechanism for improving the current agent. Conceptually, this step searches for a prompt configuration that improves expected task utility under the resource budget:
\begin{align}
P^{*}(C) = \arg\max_{P} \;\mathbb{E}_{\tau \sim \mathcal{T}}\left[
  U(\tau, A(\tau; M_\theta, P, C)) \right]
  \quad\text{s.t. } \mathrm{Cost}(A) \leq B .
\label{eq:inner}
\end{align}
This prompt-evolution tool operates entirely in textual space \citep{pryzant2023automatic,opsahl2024optimizing}. At each iteration, prompt candidates are produced through three complementary channels: \emph{(i)}~a library of handcrafted mutations that perturb the role description, reasoning directives, or sampling temperature; \emph{(ii)}~\emph{reflective} candidates, where the SLM first diagnoses each failure with a one-sentence root-cause analysis (e.g., factual confusion, logical gap, or format error) and a separate proposer call uses these diagnoses to generate targeted prompt revisions; and \emph{(iii)}~\emph{crossover} candidates, in which the two most complementary configurations on the current Pareto front are combined by asking the SLM to unify their respective strengths into a single prompt.

Candidates are evaluated on a small training mini-batch (sampled) and maintained on a Pareto front trading off accuracy against token cost. Rather than always mutating the single highest-accuracy parent, the next parent is sampled from the front with probability proportional to its failure coverage, encouraging exploration of diverse error modes. After all iterations, every front member is re-evaluated on a held-out validation split; the configuration with the best validation accuracy is retained.

Because prompt edits are low-cost, reversible, and stable under limited model capacity, PACE invokes this tool frequently under fixed control logic. Prompt evolution therefore serves as the default adaptation mode before the controller considers higher-risk structural changes.

\textbf{Constrained Control Logic Evolution (CE).}
When prompt-level gains saturate, PACE invokes control logic evolution as a higher-risk adaptation tool for modifying the agent's $C$. Conceptually, this step searches over a restricted structural space $\mathcal{C}$: $\max_{C \in \mathcal{C}}\;\mathbb{E}_{\tau \sim \mathcal{T}}\left[
  U(\tau, A(\tau; M_\theta, P^{*}(C), C))\right],$
where $\mathcal{C}$ denotes the set of admissible control-logic modifications. In practice, control-logic updates are implemented as bounded edits to the agent's solver function, which specifies how the frozen SLM is prompted, sampled, parsed, verified, or re-executed at inference time. These edits are constrained by a fixed solver interface and resource budget, and typically affect safe-to-edit components such as output parsing and validation, retry mechanisms, inference-time configuration, self-consistency or verification passes, and lightweight routing or fallback strategies.

More specifically, PACE guides solver edits using a compact failure summary over recent errors, including extraction/runtime failures, format or constraint violations, and reasoning/content mistakes. The summary steers the next proposal toward an appropriate control-logic change, such as answer-extraction hardening, validation or retry logic, sampling adjustment, self-consistency, or verification/repair. Each candidate solver edit is treated as a proposal and is committed only if it improves held-out performance under the resource budget~$B$. After an accepted update, PACE re-invokes prompt evolution to adapt $P$ to the new solver.

\subsection{Empirical Findings for Reliable SLM Agent Evolution}
Our empirical analysis suggests that reliable self-evolution under frozen SLMs depends on two practical principles. First, the agentic controller must correctly assign improvement opportunities between low-risk prompt refinement and higher-risk solver modification. Second, because SLM-generated control-logic edits can introduce subtle but severe regressions, candidate solver updates should be validated empirically before being committed. We discuss both findings below.

\subsubsection{Evolution Credit Assignment}

\textbf{Observation.}
Under frozen SLMs, prompt evolution often saturates before all error modes are resolved. Further prompt refinement under a fixed control logic $C$ may yield only marginal gains, while admissible solver modifications can still provide non-trivial improvement. The challenge is therefore deciding \emph{which adaptation tool} to invoke when performance stalls: premature control-logic edits are unstable, but excessive prompt updates waste budget once prompt-level gains have saturated.

To describe this tradeoff, we define two conceptual utility gains:
\begin{align*}
  \Delta U_P(C)
  &= \max_{P}\;
    \mathbb{E}_{\tau \sim \mathcal{T}}
    \left[ U(\tau, A(\tau; M_\theta, P, C)) \right]
  - \mathbb{E}_{\tau \sim \mathcal{T}}
    \left[ U(\tau, A(\tau; M_\theta, P_0, C)) \right], \\
  \Delta U_C(P)
  &= \max_{C' \in \mathcal{C}}\;
    \mathbb{E}_{\tau \sim \mathcal{T}}
    \left[ U(\tau, A(\tau; M_\theta, P, C')) \right]
  - \mathbb{E}_{\tau \sim \mathcal{T}}
    \left[ U(\tau, A(\tau; M_\theta, P, C)) \right],
\end{align*}
where $P_0$ is the baseline prompt and $\mathcal{C}$ is the restricted space of admissible solver modifications. These quantities are not solved exactly; they only distinguish improvements achievable through prompt refinement from those requiring changes to the inference procedure.

In practice, PACE uses recent validation improvement from the prompt-evolution tool as a conservative trigger, computed after round $t$ as $\widehat{\Delta U}_P^{(t)}=\widehat{U}(A(P_t,C);\mathcal{V})-\widehat{U}(A(P_{t-1},C);\mathcal{V})$. The controller continues prompt evolution while $\widehat{\Delta U}_P^{(t)}>\epsilon$; once $\widehat{\Delta U}_P^{(t)}\leq\epsilon$, prompt refinement is treated as saturated, and admissible control-logic proposals are considered.

\subsubsection{Empirical Validation of Solver Updates}
\textbf{Observation.}
Control-logic edits proposed by SLMs can easily lead to failures. Beyond producing invalid tool calls or non-executable code, SLMs may introduce semantically incorrect solver modifications that preserve syntactic validity but degrade performance. For instance, we observed cases where an SLM rewrote a correct solver into a verification-style procedure that omitted essential task context: the modified solver remained executable, yet caused a clear accuracy regression. Since such edits can pass runtime checks and may appear locally plausible, solver updates should not be accepted based on model self-assessment alone.

PACE addresses this by treating every control-logic edit as a candidate requiring empirical validation. We introduce
\texttt{action\_compare\_variants}, which compares two control logic variants on the same validation subset, returning per-variant performance, sample-level differences, and an accept/reject recommendation. Formally, let $A_{\mathrm{old}}$ and $A_{\mathrm{new}}$ denote the current and candidate agents. Given a validation subset $\mathcal{V} \subset \mathcal{T}$, the candidate is accepted only if
\begin{equation*}
  \widehat{U}(A_{\mathrm{new}}; \mathcal{V})
    > \widehat{U}(A_{\mathrm{old}}; \mathcal{V}) + \delta
  \quad \text{and} \quad
  \mathrm{Cost}(A_{\mathrm{new}}) \leq B,
\end{equation*}
where $\widehat{U}(\cdot; \mathcal{V})$ denotes empirical utility on the shared validation subset and $\delta \geq 0$ is the \emph{validation gate threshold}, a minimum improvement margin that filters out noisy or marginal gains. Setting $\delta{=}0$ accepts any non-regressing candidate; larger values require stronger evidence before committing the update. In practice, we set $\delta{=}0.02$, and we ensure validation comparison samples a fresh subset from the training pool, varying both size and composition across iterations. This prevents the evolution trajectory from overfitting to a fixed held-out slice and ensures that accepted control-logic updates generalize beyond any single validation draw.

\section{Experiment}
\label{sec:experiment}
We evaluate PACE on four static benchmarks using three frozen SLM backbones from 4B to 14B parameters, comparing against prompt-only and control-logic-only evolution while analyzing threshold sensitivity and token overhead. We further evaluate PACE on $\tau$-bench as a realistic multi-turn tool-use case study. Additional results and evolved solver listings are provided in the Appendix.

\subsection{Setup}
\label{sec:setup}
In our implementation, $B$ denotes the evolution budget. Each run uses a frozen SLM backbone with at most 14B parameters and is capped at 20 evolution steps. We do not enforce a fixed per-query token budget during evolution; instead, we measure the inference-time cost of the evolved agent, including model calls and generated tokens, and report the resulting accuracy--cost tradeoff (Appendix \ref{sec:token-analysis}).

\textbf{Benchmarks and Evaluation Metrics.}
We conduct controlled comparisons on four static benchmarks: MMLU~\citep{hendrycks2020measuring} for knowledge-intensive multiple-choice reasoning, MGSM~\citep{shi2022language} for multilingual math reasoning, HotpotQA~\citep{yang2018hotpotqa} for multi-hop question answering, and IFEval~\citep{zhou2023instruction} for verifiable instruction following. We report task-standard accuracy metrics, including letter-match accuracy for MMLU, exact numeric match for MGSM, exact match for HotpotQA, and strict instruction accuracy for IFEval. We further evaluate PACE on $\tau$-bench~\citep{yao2024tau} as a realistic multi-turn tool-use case study involving simulated users, domain policies, and API tools. Full dataset splits and evaluation details are in Appendix~\ref{sec:benchmark-details}. Note that the utility $U$ is instantiated by each benchmark's primary metric: accuracy for MMLU, exact match for MGSM and HotpotQA, strict accuracy for IFEval, and task success for $\tau$-bench.

\textbf{SLM backbones.}
All experiments use frozen SLMs served locally via vLLM~\citep{kwon2023efficient}. We study three backbone families ranging from 4B to 14B parameters:
1)~\textsc{Qwen/Qwen3-4B-Instruct-2507}~\citep{yang2025qwen3},
2)~\textsc{Qwen/Qwen3.5-9B}, and
3)~\textsc{Ministral-3-14B-Instruct-2512}. Each backbone serves both as the inference model inside the solver and as the controller model that proposes control-logic edits. No stronger teacher model is used during evolution.

\begin{table}[t!]
\centering
\small
\caption{Performance comparison, where CE denotes control logic evolution only, PE denotes prompt evolution only, and $+\%$ denotes relative performance gain. Best result per backbone is \textbf{bolded}.}
\label{tab:main_results}
\resizebox{0.85\textwidth}{!}{
\begin{tabular}{llcccccccc}
\toprule
\textbf{SLM} & \textbf{Method}
& \textbf{MMLU} & \textbf{+\%}
& \textbf{MGSM} & \textbf{+\%}
& \textbf{HotpotQA} & \textbf{+\%}
& \textbf{IFEval} & \textbf{+\%} \\
\midrule
& DeepSeek V3.2
& 0.908 & {--}
& 0.932 & {--}
& 0.779 & {--}
& 0.865 & {--} \\
\midrule

\multirow{8}{*}{\rotatebox{90}{Qwen3-4B}}
& Vanilla
& 0.771 & {--}
& 0.795 & {--}
& 0.775 & {--}
& 0.691 & {--} \\
\cmidrule(l){2-10}
& MIPROv2
& 0.771 & +0.0
& 0.795 & +0.0
& 0.758 & -2.1  
& 0.691 & +0.0 \\
& GEPA
& 0.771 & +0.0
& 0.802 & +0.1
& 0.775 & +0.0
& 0.676 & -0.1 \\
& ACE
& 0.800 & +3.7
& 0.808 & +1.6
& 0.775 & +0.0
& 0.711 & +2.9 \\
& G\"{o}del Agent
& 0.797 & +3.3
& 0.812 & +2.1
& 0.772 & -0.5
& 0.705 & +2.0 \\
\cmidrule(l){2-10}
& +CE
& 0.797 & +3.3
& 0.821 & +3.3
& 0.770 & -0.1
& 0.722 & +4.4 \\
& +PE
& 0.788 & +2.2
& 0.815 & +2.5
& 0.775 & +0.0  
& 0.705 & +2.0 \\
& \textbf{+PACE}
& \textbf{0.818} & \textbf{+6.1}
& \textbf{0.831} & \textbf{+4.3}
& \textbf{0.776} & +0.0
& \textbf{0.727} & \textbf{+5.2} \\
\midrule

\multirow{8}{*}{\rotatebox{90}{Qwen3.5-9B}}
& Vanilla
& 0.818 & {--}
& 0.858 & {--}
& 0.754 & {--}
& 0.697 & {--} \\
\cmidrule(l){2-10}
& MIPROv2
& 0.797 & -2.5
& 0.861 & +0.4
& 0.739 & -2.0
& 0.697 & +0.0 \\
& GEPA
& 0.832 & +1.7
& 0.878 & +2.3
& 0.741 & -1.7
& 0.704 & +0.1 \\
& ACE
& 0.805 & -1.5
& 0.889 & +3.6
& 0.771 & +2.3
& 0.718 & +3.0 \\
& G\"{o}del Agent
& 0.835 & +2.1
& 0.864 & +0.7
& 0.780 & +3.5
& 0.732 & +5.0 \\
\cmidrule(l){2-10}
& +CE
& 0.835 & +2.1
& 0.904 & +5.3
& 0.752 & -0.2
& 0.725 & +4.0 \\
& +PE
& 0.832 & +1.7
& 0.872 & +1.6
& 0.789 & +4.6
& 0.718 & +3.0 \\
& \textbf{+PACE}
& \textbf{0.889} & \textbf{+8.6}
& \textbf{0.909} & +\textbf{5.9}
& \textbf{0.803} & +\textbf{6.5}
& \textbf{0.761} & \textbf{+9.2} \\
\midrule

\multirow{8}{*}{\rotatebox{90}{Ministral-3-14B}}
& Vanilla
& 0.800 & {--}
& 0.770 & {--}
& 0.596 & {--}
& 0.781 & {--} \\
\cmidrule(l){2-10}
& MIPROv2
& 0.833 & +4.1
& 0.783 & +1.6
& 0.583 & -2.2
& 0.801 & +2.6 \\
& GEPA
& 0.825 & +3.1
& 0.775 & +0.6
& 0.625 & +4.8
& 0.816 & +4.5 \\
& ACE
& 0.818 & +2.2
& 0.785 & +1.9
& 0.605 & +1.5
& 0.788 & +0.1 \\
& G\"{o}del Agent
& 0.797 & +0.0
& 0.825 & +7.1
& 0.659 & +10.6
& 0.810 & +3.7 \\
\cmidrule(l){2-10}
& +CE
& 0.836 & +4.5
& 0.880 & +14.3
& 0.654 & +9.7
& 0.828 & +6.0 \\
& +PE
& 0.832 & +4.4
& 0.814 & +5.5
& 0.709 & +18.9
& 0.816 & +4.5 \\
& \textbf{+PACE}
& \textbf{0.854} & +\textbf{6.7}
& \textbf{0.908} & +\textbf{17.9}
& \textbf{0.774} & +\textbf{29.8}
& \textbf{0.837} & \textbf{+7.2} \\
\bottomrule
\end{tabular}}
\vspace{-7mm}
\end{table}


\textbf{Baselines.} We compare PACE against the following: 1. \textit{Vanilla SLM}: the unmodified solver with a task-specific prompt and default temperature ($T{=}0.2$). 2. \textit{+PE} (prompt evolution only): we run prompt updates to convergence (starts from a minimal and unified template); no CE are permitted. 3. \textit{+CE} (control logic evolution only): the agent may modify control logic from the start, without a dedicated prompt optimization phase. 4. \textit{GEPA}~\citep{agrawal2025gepa}: declarative prompt compilation pipeline. 5.  \textit{MIPROv2}~\citep{opsahl2024optimizing}: multi-stage instruction and demonstration optimizer. 6. \textit{G\"{o}del Agent}~\citep{yin2025godel}: a self-referential agent framework for recursive control-logic updates. 7. \textit{ACE}~\citep{zhang2025agentic}: an agent self-evolution by using execution feedback to iteratively reflect on successes and failures, curate reusable lessons into a structured context/playbook, and improve future task performance without updating model weights. For a frontier reference, we also report \texttt{DeepSeek-V3.2-685B} results on each benchmark. Note that all experiments are conducted over $10$ independent runs with different random seeds, and we report the mean accuracy in the main context. Results with standard deviation are reported in Table \ref{tab:main_results_std} in Appendix.

\subsection{Quantitative Results}
Table~\ref{tab:main_results} reports performance across three SLM backbones on four controlled benchmarks, with DeepSeek-V3.2 included as a frontier reference. Across all SLM backbones, PACE achieves the best overall performance among the evaluated adaptation methods. The strongest gains are observed on MMLU (+$6.1\%$ to +$8.6\%$ accuracy) and IFEval (+$5.2\%$ to +$9.2\%$), where PACE consistently outperforms both prompt-only (+PE) and control-logic-only (+CE) evolution. Compared with the DeepSeek-V3.2 reference, PACE substantially narrows the vanilla-to-frontier gap. For example, on Qwen3.5-9B, PACE closes $79\%$ of the MMLU gap and $38\%$ of the IFEval gap, while on Ministral-3-14B it closes 50\% of the MMLU gap and $67\%$ of the IFEval gap.

\textbf{Comparison to single-axis evolution.}
Across backbones, +PE and +CE each capture partial gains, while PACE combines and exceeds them. For example, on Qwen3.5-9B MMLU, +PE reaches 0.832 and +CE reaches $0.835$, whereas PACE achieves $0.889$, approaching the DeepSeek-V3.2 reference score of 0.908. On Qwen3.5-9B IFEval, PACE improves by +$9.2\%$ over vanilla, compared with +$3.0\%$ for +PE and +$4.0\%$ for +CE. These results suggest that prompt and control-logic updates address complementary failure modes. The evolution trajectories in Figure~\ref{fig:trajectory} further illustrate why both mechanisms are needed: prompt-only evolution saturates, control-logic-first evolution is noisy, and PACE delays structural updates until prompt gains plateau.

\textbf{Comparison to existing optimizers.}
Prompt/context optimizers such as MIPROv2, GEPA, and ACE show inconsistent gains and occasionally regress, reflecting their inability to repair structural bottlenecks such as brittle parsing, missing validation, or weak retry logic. G\"{o}del Agent, which permits structural modification, is more competitive but still trails PACE on most benchmarks. We attribute this gap to PACE's saturation-gated transition and validation-gated acceptance, which reduce premature or harmful control-logic edits under SLM constraints.

\textbf{Benchmark-specific trends.}
On MGSM and HotpotQA, +CE already provides large gains for the 9B and 14B backbones, and PACE further improves performance, indicating that multi-step reasoning and retrieval-style tasks benefit strongly from CE such as verification, structured reasoning passes, or retry mechanisms. Notably, PACE with Qwen3.5-9B surpasses the DeepSeek-V3.2 reference on HotpotQA, reaching $0.803$ compared with $0.779$, while Ministral-3-14B with PACE nearly matches DeepSeek-V3.2 on HotpotQA, reaching $0.774$. In contrast, Qwen3-4B shows little improvement on HotpotQA across all methods, suggesting that model capacity can remain the limiting factor even with improved agent design. On MMLU, prompt evolution contributes a larger share of the gain, consistent with the importance of improved elicitation for knowledge-intensive tasks.

\textbf{Cost analysis.}
PACE incurs a one-time evolution cost of approximately 2--3M generated tokens for Qwen3.5-9B, depending on the benchmark. At inference time, PE adds little overhead, typically 1.1--1.2$\times$ the baseline token cost, while accepted CE increases cost by introducing additional sampling, verification, or repair calls. For example, the evolved MMLU solver uses about 3.6$\times$ the baseline per-query tokens, while the evolved IFEval solver uses about 5.1$\times$. These results show that PACE trades additional inference-time compute for accuracy, and that the cost is concentrated in structural updates rather than prompt refinement. Detailed token breakdowns are provided in Appendix~\ref{sec:token-analysis}.

\begin{table}[t!]
\centering
\small
\caption{Performance on $\tau$-bench\tablefootnote{We omit the CE-only baseline because multi-turn dialogue performance is highly sensitive to prompt quality; applying structural evolution from an unoptimized prompt yields unstable trajectories that do not constitute a meaningful ablation.}: we leverage Claude Sonnet 4.5 as the stochastic user simulator across Retail (115 tasks) and Airline (50 tasks) domains. $\text{pass}^k$ denotes the fraction of tasks solved in \emph{all} $k$ independent trials. Best result per backbone is \textbf{bolded}.}
\label{tab:tau_bench}
\setlength{\tabcolsep}{4pt}
\resizebox{0.85\textwidth}{!}{
\begin{tabular}{@{}llcccccccc@{}}
\toprule
& & \multicolumn{4}{c}{\textbf{Retail} (115 tasks)}
& \multicolumn{4}{c}{\textbf{Airline} (50 tasks)} \\
\cmidrule(lr){3-6} \cmidrule(lr){7-10}
\textbf{SLM} & \textbf{Method}
& pass\textasciicircum1 & pass\textasciicircum2 & pass\textasciicircum3 & pass\textasciicircum4
& pass\textasciicircum1 & pass\textasciicircum2 & pass\textasciicircum3 & pass\textasciicircum4 \\
\midrule
\rowcolor{gray!10}
& Sonnet 3.5 \citep{yao2024tau}
& .692 & .576 & .509 & .462
& .460 & .326 & .263 & .225 \\

\rowcolor{gray!10}
& GPT-4o \citep{yao2024tau}
& .604 & .491 & .430 & .383
& .420 & .273 & .220 & .200\\

\midrule
\multirow{3}{*}{Qwen3-4B}
& Vanilla
& .465 & .332 & .261 & .217
& .265 & \textbf{.160} & .130 & .120 \\
& +PE
& .474 & .341 & .278 & .226
& .235 & .097 & .050 & .040 \\
& \textbf{+PACE}
& \textbf{.513} & \textbf{.378} & \textbf{.307} & \textbf{.270}
& \textbf{.275} & .153 & \textbf{.100} & \textbf{.060} \\
\midrule
\multirow{3}{*}{Qwen3.5-9B}
& Vanilla
& .791 & .706 & .659 & .626
& .535 & .417 & .355 & .320 \\
& +PE
& .785 & .704 & .652 & .617
& .550 & .450 & .390 & .340 \\
& \textbf{+PACE}
& \textbf{.817} & \textbf{.745} & \textbf{.698} & \textbf{.661}
& \textbf{.575} & \textbf{.457} & \textbf{.395} & \textbf{.360} \\
\bottomrule
\end{tabular}}
\vspace{-7mm}
\end{table}


\textbf{$\tau$-bench case study.}
Table~\ref{tab:tau_bench} evaluates PACE on realistic multi-turn tool-use tasks, where the agent must maintain dialogue state, follow domain policies, and issue valid API calls while interacting with a stochastic user simulator.
Prompt-only evolution is unstable in this setting: +PE improves Qwen3-4B Retail but degrades Qwen3.5-9B Retail ($0.785$ vs.\ $0.791$) and collapses on Qwen3-4B Airline ($\text{pass}^4$: $0.040$ vs.\ $0.120$).
PACE stabilizes evolution and achieves the best results across all configurations, with gains amplifying at higher $\text{pass}^k$ ($+3.5$~pp at $\text{pass}^4$ vs.\ $+2.6$~pp at $\text{pass}^1$ on Qwen3.5-9B Retail), indicating improved behavioral consistency across repeated user interactions rather than single-trial luck.
Notably, Qwen3.5-9B + PACE surpasses both the Sonnet~3.5 and GPT-4o references~\citep{yao2024tau} on Retail and Airline across all $\text{pass}^k$ levels.


\subsection{Ablation Study and Parameter Sensitivity Analysis}
\label{sec:ablation}

We ablate two core design choices in PACE using Qwen3.5-9B on MMLU and IFEval: (i)~the credit-assignment threshold~$\varepsilon$ that gates the transition from prompt optimization to structural exploration, and (ii)~the validation gate threshold~$\delta$ that filters proposed solver edits before they are committed. Table~\ref{tab:ablation} reports test accuracy under each setting. For the $\varepsilon$ sweep (left), $\varepsilon{=}0$ disables control logic evolution entirely (prompt-only), while $\varepsilon{=}1.0$ bypasses saturation detection and unlocks structure edits after the first prompt round.  For the $\delta$ sweep (right), $\delta{=}{-}1$ disables the gate (any candidate is accepted), while $\delta{=}0.05$ requires the candidate to exceed the current solver by at least 5\%.

\begin{table}[t]
\centering
\caption{Ablation on credit-assignment threshold $\varepsilon$ (left) and control logic validation gate $\delta$ (right) with Qwen 3.5-9B as the base model.  Shaded columns mark the proposed defaults ($\varepsilon{=}0.01$, $\delta{=}0.02$).}
\label{tab:ablation}
\small
\setlength{\tabcolsep}{4pt}
\begin{tabular}{l cccccc c cccc}
\toprule
& \multicolumn{6}{c}{Credit-Assignment Threshold ($\varepsilon$)}
& \phantom{a}
& \multicolumn{4}{c}{Validation Gate Threshold ($\delta$)} \\
\cmidrule(lr){2-7} \cmidrule(lr){9-12}
Benchmark
  & 0 & .005 & \cellcolor{gray!15}.01 & .02 & .05 & 1.0
  &
  & $-$1 & 0 & \cellcolor{gray!15}.02 & .05 \\
\midrule
MMLU
  & 83.2 & 87.9 & \cellcolor{gray!15}\textbf{88.9} & 88.6 & 88.3 & 88.4
  &
  & 88.3 & 88.8 & \cellcolor{gray!15}\textbf{88.9} & 88.9 \\
IFEval
  & 71.8 & 73.4 & \cellcolor{gray!15}\textbf{76.1} & 74.8 & 73.1 & 72.5
  &
  & 72.3 & 74.6 & \cellcolor{gray!15}\textbf{76.1} & 75.2 \\
\bottomrule
\end{tabular}
\vspace{-6mm}
\end{table}

\textbf{Credit-assignment threshold~$\varepsilon$.} In Table~\ref{tab:ablation}, both extremes hurt: $\varepsilon{=}0$ (prompt-only) never enters structural search, while $\varepsilon{=}1.0$ unlocks structure edits before the prompt is sufficiently refined. The default $\epsilon=0.01$ achieves the best accuracy on both benchmarks, improving over the prompt-only setting by $\sim5\%$ on both benchmarks. These gains indicate CE can provide substantial improvements beyond PE alone. The $\varepsilon=1.0$ setting approximates a naïve combination that allows control-logic edits before prompt refinement has saturated. Its lower performance, especially on IFEval, shows combining prompt and control-logic search is insufficient; the transition policy matters.

\textbf{Validation gate threshold~$\delta$.} Disabling the gate ($\delta{=}{-}1$) degrades IFEval by $3.8\%$, as harmful edits pass unchecked. On MMLU the effect is smaller, since MMLU's multiple-choice format leaves less room for structurally broken solvers to diverge. Overly strict gating ($\delta{=}0.05$) also underperforms the default, as it blocks beneficial edits that show moderate but genuine improvement on the small validation subset. The default $\delta{=}0.02$ strikes the best balance, filtering noisy candidates while admitting meaningful structural gains.





\subsection{Failure Mode Shift Across Evolution Phases}
\label{sec:failure-shift}

The two-timescale design rests on the premise that prompt and structural interventions resolve \emph{different} failure modes. To validate this, we classify every failure on a 20-sample mini-batch into three categories---\emph{extraction/runtime} (malformed output, parse errors), \emph{format/constraint} (valid output violating task requirements), and \emph{reasoning/content} (well-formed but incorrect)---and track how the distribution shifts from Vanilla through +PE to +PACE.

\begin{figure}[t]
\centering
\includegraphics[width=\linewidth]{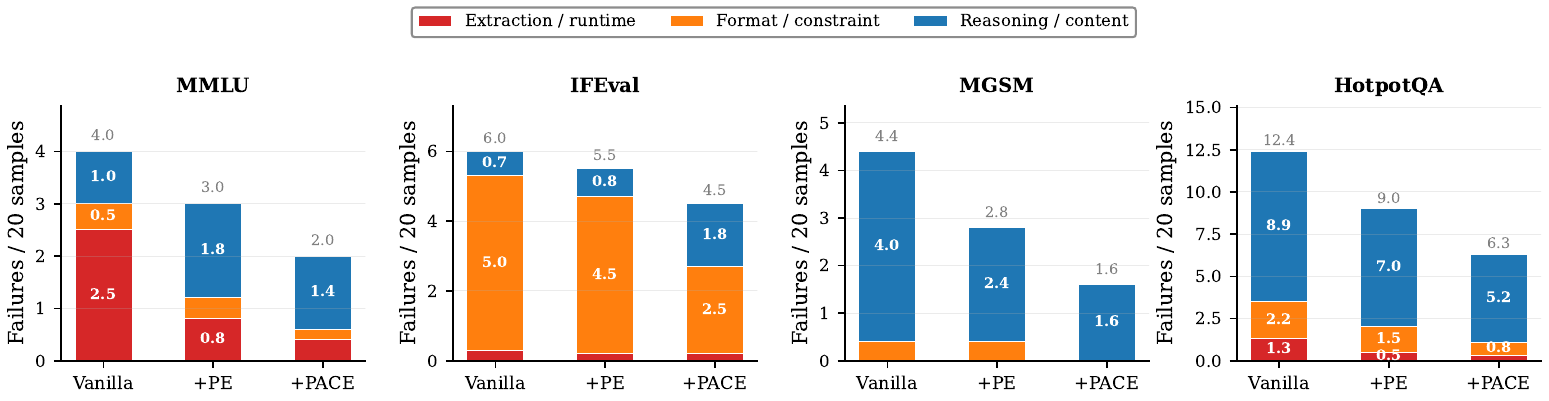}
\vspace{-4mm}
\caption{Failure mode distribution per 20-sample mini-batch across evolution phases on all benchmarks  with Qwen3.5-9B. PE resolves extraction and format errors; residual failures after +PE are dominated by reasoning or hard constraint violations, which CE then targets.}
\label{fig:failure_shift}
\vspace{-5mm}
\end{figure}

Fig.~\ref{fig:failure_shift} further shows PE reduces extraction/format errors, while the residual failures after +PE are increasingly dominated by reasoning/content or hard-constraint errors. This shift supports the saturation trigger: once prompt-level gains plateau, the remaining errors are more likely to require changes to the solver's inference procedure, such as voting, verification, or task-specific extraction.

\subsection{Control-Logic Proposal Filtering}\vspace{-2mm}
\label{sec:proposal-filtering}
\begin{table}[t]
\centering
\small
\caption{Control-logic proposal filtering during PACE evolution, aggregated over three runs (with random seeds) per task. Each entry reports Qwen3-4B / Qwen3.5-9B. ``Rej.\ regression'' denotes rejected executable candidates with non-positive validation gain.}
\label{tab:proposal-filtering}
\resizebox{0.75\linewidth}{!}{
\begin{tabular}{l cc cc cc cc cc}
\toprule
\textbf{Task}
 & \multicolumn{2}{c}{\textbf{Proposed}}
 & \multicolumn{2}{c}{\textbf{Executable}}
 & \multicolumn{2}{c}{\textbf{Accepted}}
 & \multicolumn{2}{c}{\textbf{Rejected}}
 & \multicolumn{2}{c}{\textbf{Rej.\ regression}} \\
\cmidrule(lr){2-3} \cmidrule(lr){4-5} \cmidrule(lr){6-7} \cmidrule(lr){8-9} \cmidrule(lr){10-11}
 & 4B & 9B & 4B & 9B & 4B & 9B & 4B & 9B & 4B & 9B \\
\midrule
MMLU     & 31 & 38 & 26 & 36 & 19 & 28 &  7 &  8 &  7 &  8 \\
MGSM     & 41 & 47 & 34 & 43 & 22 & 29 & 12 & 14 & 11 & 13 \\
HotpotQA & 44 & 52 & 37 & 46 & 22 & 28 & 15 & 18 & 14 & 17 \\
IFEval   & 24 & 29 & 21 & 28 & 17 & 22 &  4 &  6 &  4 &  6 \\
\bottomrule
\end{tabular}}
\vspace{-4mm}
\end{table}
To quantify the role of the validation gate, we track all control-logic proposals generated by \texttt{action\_adjust\_logic} across four benchmarks and two backbones. For each proposal, we record whether it is executable, accepted by held-out validation, or rejected despite being executable. Table~\ref{tab:proposal-filtering} shows that frozen SLMs are useful but noisy generators of solver edits. The 4B model is less stable than the 9B model, producing fewer executable and accepted proposals, but held-out validation filters non-improving edits for both backbones. Nearly all rejected executable edits have non-positive validation gain, supporting PACE's proposal--validation separation: the SLM proposes candidate control-logic changes, while validation determines which updates are committed.

\vspace{-1mm}
\section{Conclusion}\vspace{-2mm}
We introduced PACE, a two-timescale agentic framework for frozen SLM self-evolution that coordinates fast prompt refinement with less frequent, constrained control-logic updates. By separating proposal from validation, PACE lets the SLM generate candidate edits while committing only those that improve held-out validation performance under a resource budget. Across all benchmarks and three SLM backbones (4B--14B), PACE outperforms prompt-only and control-logic-only evolution, achieving up to +$9.2\%$ accuracy over vanilla agents at a one-time evolution cost of 2--3M tokens. More broadly, PACE shows that the novelty is not in any single final solver pattern, but in enabling a frozen SLM to autonomously discover, select, and validate task-appropriate inference strategies; the resulting accepted and rejected evolution trajectories also provide preference- or reward-labeled data for training future evolution controllers.

\bibliographystyle{unsrtnat}
\bibliography{references}

\appendix
\section{Appendix}

\subsection{Benchmark Details and Data Splits}
\label{sec:benchmark-details}

This section documents the four benchmarks used throughout the paper
(MMLU, IFEval, HotpotQA, and MGSM) and specifies the data splits,
program structure, and evaluation metrics shared by the Vanilla,
+PE, and +PACE conditions as well as the DSPy baselines (GEPA,
MIPROv2) reported in the main results.

\paragraph{Common protocol.}
Every benchmark is partitioned into three disjoint, seed-shuffled
splits.  The \emph{train} split is used by the inner-loop prompt
optimizer to sample minibatches and by DSPy baselines to bootstrap
demonstrations.  The \emph{validation} split is used for
comparisons of candidate solvers or prompts (including the PACE
credit-assignment gate).  The \emph{test} split is held out and used
only for the final reported numbers.  All experiments use seed $0$,
the \texttt{auto="medium"} budget preset for DSPy optimizers, and
identical splits across Vanilla, +PE, +PACE, GEPA, and MIPROv2 so
that differences reflect the method rather than the data.

Each task is implemented as a single-module DSPy \texttt{ChainOfThought}
predictor.  The signature exposes an input schema, a
\texttt{reasoning} field, and a task-specific output field; only the
signature instruction is subject to optimization.  At inference time
the model is served via an OpenAI-compatible interface backed by
vLLM, with a round-robin dispatcher sharding requests across
replicas.

\begin{table}[h]
\centering
\small
\caption{Controlled data splits used for repeated evolution experiments. All methods use the same train/test subsets; validation subsets are dynamically sampled by the agent during evolution. Results should be interpreted as within-protocol comparisons rather than standard leaderboard scores.}
\label{tab:splits}
\setlength{\tabcolsep}{5pt}
\begin{tabular}{@{}l c c c l l@{}}
\toprule
\textbf{Task} & \textbf{Train} & \textbf{Val} & \textbf{Test}
  & \textbf{Test source} & \textbf{Primary metric} \\
\midrule
MMLU     & 200 & dynamic & 800 & official test subset        & accuracy \\
IFEval   & 200 & dynamic & 141 & held-out from official prompts & strict instruction accuracy \\
HotpotQA & 200 & dynamic & 500 & official validation subset  & exact match ($+$ F1) \\
MGSM     & 200 & dynamic & 800 & pooled multilingual subset  & exact numeric match \\
\bottomrule
\end{tabular}
\end{table}

Because evolution requires repeated evaluation, we use fixed controlled subsets for training and testing rather than full benchmark test suites; the same splits are shared across all methods. For IFEval, whose official release consists of a single set of $541$ prompts without a designated train/test partition, we randomly split the full prompt set into disjoint train and test subsets using a fixed seed.

Validation subsets are not fixed: at each evaluation step, the agent dynamically samples a fresh subset from the training pool, varying both size and composition across iterations to reduce overfitting to any single validation slice. This design reflects a practical deployment constraint where repeated evaluation on an identical held-out set can lead to implicit selection bias in the evolution trajectory.

Following standard practice in iterative prompt/agent optimization \citep{opsahl2024optimizing,pryzant2023automatic}, we evaluate on fixed controlled subsets to enable repeated evaluation during evolution while keeping compute tractable. All methods share identical splits, ensuring fair within-protocol comparison.

\paragraph{MMLU (multiple-choice knowledge).}
MMLU is a closed-book multiple-choice benchmark covering 57 subjects with four answer choices per question.
\begin{itemize}[nosep,leftmargin=*]
\item \textbf{Data.} The full dataset is loaded from Kaggle\footnote{https://www.kaggle.com/datasets/open-benchmarks/mmlu-massive-multitask-language-understanding/data}. We sampled $800$ cases as held-out test set.
\item \textbf{Program.} \texttt{ChainOfThought} over a signature with input \texttt{question} and outputs \texttt{reasoning} and \texttt{answer}, where \texttt{answer} is constrained to one letter in $\{A,B,C,D\}$.
\item \textbf{Metric.} We report letter-match accuracy. The first A/B/C/D token is extracted from the model output case-insensitively, and the prediction is counted correct iff it matches the gold letter.
\end{itemize}

\paragraph{IFEval (verifiable instruction following).}
IFEval measures whether a model satisfies verifiable formatting and content constraints attached to each prompt.
\begin{itemize}[nosep,leftmargin=*]
\item \textbf{Data.} We shuffle the available IFEval prompts and use 200 examples for training and 100 examples for validation. A disjoint 141-example subset from the official test set is reserved for final evaluation.
\item \textbf{Program.} \texttt{ChainOfThought} over a signature with input \texttt{prompt} and outputs \texttt{reasoning} and \texttt{response}. The \texttt{response} field is scored by the instruction checkers.
\item \textbf{Metrics.} We use the IFEval instruction checkers for length, case, keyword, format, and punctuation constraints. Each response is scored by \emph{strict accuracy}, which equals $1.0$ only if all attached instructions are satisfied, and \emph{loose accuracy}, the fraction of satisfied instructions. Strict accuracy is used as the optimization objective; failure feedback lists violated instruction identifiers and the loose score.
\end{itemize}

\paragraph{HotpotQA (multi-hop QA, distractor setting).}
HotpotQA~\cite{yang2018hotpotqa} requires combining evidence from multiple Wikipedia paragraphs to answer a question. In the distractor setting, each example includes ten paragraphs, consisting of two gold paragraphs and eight distractors.
\begin{itemize}[nosep,leftmargin=*]
\item \textbf{Data.} We use the HuggingFace \texttt{hotpot\_qa} distractor split. After shuffling the training split, we use 200 examples for training and 100 for validation. Final evaluation uses the first 500 examples from the official validation split. Paragraphs are concatenated into a single context string with title markers and sentence text.
\item \textbf{Program.} \texttt{ChainOfThought} over a signature with inputs \texttt{context} and \texttt{question} and outputs \texttt{reasoning} and \texttt{answer}, where \texttt{answer} is a short span.
\item \textbf{Metrics.} We apply SQuAD-style normalization, including lowercasing, article removal, punctuation removal, and whitespace normalization. Exact Match is used as the optimizer objective, and token-level F1 is also reported for analysis.
\end{itemize}

\paragraph{MGSM (multilingual grade-school math).}
MGSM is a multilingual grade-school math benchmark with numeric answers, covering 11 languages: Bengali, German, English, Spanish, French, Japanese, Russian, Swahili, Telugu, Thai, and Chinese.
\begin{itemize}[nosep,leftmargin=*]
\item \textbf{Data.} For each language, we load \texttt{mgsm\_\{lang\}.tsv}, prepend a language-specific instruction prefix, pool examples across languages, shuffle with seed $0$, and split the pooled data into 200 training, 100 validation, and 800 test examples.
\item \textbf{Program.} \texttt{ChainOfThought} over a signature with input \texttt{question} and outputs \texttt{reasoning} and \texttt{answer}, where \texttt{answer} is numeric.
\item \textbf{Metric.} We report exact numeric match. The model output is normalized by extracting the first numeric token, removing thousands separators, and trimming trailing zeros and decimal points; the prediction is correct iff the normalized value matches the gold answer.
\item \textbf{Inference note.} Non-English problems occasionally produce long repetitive generations, so we set \texttt{max\_tokens} to $8000$ and use task-generation temperature $0.6$ for MGSM. Parse failures caused by truncation are counted as incorrect rather than aborting evaluation.
\end{itemize}

\paragraph{$\tau$-bench (multi-turn tool-augmented dialogue).}
$\tau$-bench~\citep{yao2024tau} is a realistic agent benchmark that evaluates LLM agents in dynamic, multi-turn customer-service conversations requiring policy-compliant tool use. The agent interacts with a stochastic LLM-simulated user who has a hidden intent (e.g., exchange an item, cancel an order), and must resolve the request by calling domain-specific API tools (e.g., \texttt{get\_order\_details}, \texttt{exchange\_delivered\_order\_items}) while adhering to a detailed policy document (the ``wiki'').

\begin{itemize}[nosep,leftmargin=*]
\item \textbf{Data.}  We evaluate on two domains: \emph{Retail} (115 tasks involving order management, returns, exchanges, and address modifications) and \emph{Airline} (50 tasks involving flight changes, cancellations, and baggage policies). Each task specifies a user persona with a hidden intent and a gold action sequence; the environment is fully stateful with a database backend.
\item \textbf{Program.}  The agent runs a multi-turn loop: at each step it receives the user's message (or tool result), generates either a tool call or a natural-language response, and the environment advances accordingly. The system prompt consists of the domain policy wiki concatenated with any evolved role description and requirements. The agent has access to 10--15 domain-specific tools per environment. Maximum conversation length is 30 steps (tool calls + responses).

\item \textbf{Metric.}  Each task is evaluated over $k{=}4$ independent trials with the stochastic user simulator. $\text{pass}^k$ denotes the fraction of tasks where the agent succeeds (reward $= 1.0$) in \emph{all} $k$ trials, computed via the combinatorial estimator $\text{pass}^k_i = \binom{c_i}{k} / \binom{n}{k}$ where $c_i$ is the number of successful trials for task $i$.
This metric captures both accuracy and behavioral consistency under user variability.

\item \textbf{User simulator.}  We use Claude Sonnet~4.5 (\texttt{claude-sonnet-4-5-20250929}) as the user simulator for all reported results. The simulator follows the task's hidden intent, responds naturally to the agent, and terminates the conversation (\texttt{\#\#\#STOP\#\#\#}) if the agent becomes unresponsive or completes the task.

\item \textbf{Inference note.}  The agent model (Qwen3.5-9B) is served locally via vLLM on 2 A100 GPUs with temperature $0.0$. During evolution, validation uses single-trial evaluation on all tasks for efficiency; the final reported metrics use 4-trial evaluation. Each full 4-trial evaluation run takes approximately 8 hours per domain.
\end{itemize}

\paragraph{DSPy baselines.}
The GEPA and MIPROv2 baselines reported in the main tables use the splits above, the same \texttt{ChainOfThought} program for each task, and identical evaluation metrics.  Both optimizers are run with \texttt{auto="medium"} and seed $0$.  GEPA additionally uses a reflection language model at temperature $1.0$; MIPROv2 uses its default grounded proposer.  To keep comparisons fair with PACE, DSPy's \texttt{max\_errors} threshold is set to the size of the evaluation set so that sporadic parse failures on pathological examples are counted as incorrect rather than aborting the evaluation.

\paragraph{Inference infrastructure.}
All experiments run on a single node with eight vLLM replicas of the target SLM, each bound to one GPU.  When four experiments share the cluster we allocate two replicas per task and use $24$ evaluation threads; when a single task has exclusive use of the cluster we use all eight replicas with $48$ threads.  Task-generation temperature is $0.2$ for MMLU, IFEval, and HotpotQA and $0.6$ for MGSM; optimizer-side reflection temperature is fixed at $1.0$.

\subsection{Experiment Settings}
\textbf{Agent Tools Description.} The following list covers existing tools for agent control logic (i.e., outer loop) optimization. Note that \texttt{action\_optimize\_prompt\_on\_task} is the inner loop prompt optimization that would run by default in each iteration.
\begin{itemize}[leftmargin=*]
    \item \texttt{action\_display\_analysis}: Summarize the latest evaluation failures into a structured failure taxonomy (extraction/runtime, format/constraint, reasoning/content) and produce actionable evolution guidance with ranked failure modes and fix recommendations.
    \item \texttt{action\_read\_logic}: Read the source code of a specified function, method, or class within a given module.
    \item \texttt{action\_adjust\_logic}: Modify, add, or delete the source code of a specified function, method, or class within a given module to improve task-solving ability.
    \item \texttt{action\_run\_code}: Execute Python or shell code and capture the output, errors, and return value.
    \item \texttt{action\_call\_json\_format\_llm}: Call an external LLM for assistance with gathering insights, refining strategies, correcting errors, and solving complex problems, returning the response in JSON format.
    \item \texttt{action\_select\_examples}: Select a representative subset of task examples (diverse, random, or head strategy) for future train, valid, or evaluate calls.
    \item \texttt{action\_compare\_variants}: Run a candidate comparison of two solver or prompt variants on the same sampled validation subset to verify whether a change improves accuracy.
    \item \texttt{action\_optimize\_prompt\_on\_task}: Run the inner-loop prompt optimizer, evolving role, requirements, and temperature using minibatch feedback.
    \item \texttt{action\_get\_evolution\_credit}: Return the current evolution credit assignment state, including $\Delta U_P$, $\Delta U_C$, prompt saturation status, and thresholds.
    \item \texttt{action\_evaluate\_on\_task}: Evaluate the current solver on the goal task samples and return evaluation feedback including accuracy and per-sample details.
\end{itemize}

\begin{table}[t!]
\centering
\small
\caption{Performance comparison across SLM backbones. CE denotes control-logic evolution only, PE denotes prompt evolution only, and PACE denotes the full two-timescale framework. Results are reported as mean $\pm$ standard deviation over multiple runs. Best result per backbone is \textbf{bolded}.}
\label{tab:main_results_std}
\resizebox{0.85\textwidth}{!}{
\begin{tabular}{llcccc}
\toprule
\textbf{SLM} & \textbf{Method}
& \textbf{MMLU}
& \textbf{MGSM}
& \textbf{HotpotQA}
& \textbf{IFEval} \\
\midrule
\multirow{8}{*}{\rotatebox{90}{Qwen3-4B}}
& Vanilla
& $0.771 \pm 0.005$
& $0.795 \pm 0.006$
& $0.775 \pm 0.012$
& $0.691 \pm 0.008$ \\
\cmidrule(l){2-6}
& MIPROv2
& $0.771 \pm 0.005$
& $0.795 \pm 0.007$
& $0.758 \pm 0.014$
& $0.691 \pm 0.009$ \\
& GEPA
& $0.771 \pm 0.005$
& $0.802 \pm 0.007$
& $0.775 \pm 0.013$
& $0.676 \pm 0.010$ \\
& ACE
& $0.800 \pm 0.006$
& $0.808 \pm 0.008$
& $0.775 \pm 0.015$
& $0.711 \pm 0.011$ \\
& G\"{o}del Agent
& $0.797 \pm 0.015$
& $0.812 \pm 0.022$
& $0.772 \pm 0.048$
& $0.705 \pm 0.055$ \\
\cmidrule(l){2-6}
& +CE
& $0.797 \pm 0.021$
& $0.821 \pm 0.024$
& $0.770 \pm 0.038$
& $0.722 \pm 0.031$ \\
& +PE
& $0.788 \pm 0.006$
& $0.815 \pm 0.008$
& $0.775 \pm 0.014$
& $0.705 \pm 0.010$ \\
& \textbf{+PACE}
& $\mathbf{0.818 \pm 0.007}$
& $\mathbf{0.831 \pm 0.009}$
& $\mathbf{0.776 \pm 0.015}$
& $\mathbf{0.727 \pm 0.011}$ \\
\midrule
\multirow{8}{*}{\rotatebox{90}{Qwen3.5-9B}}
& Vanilla
& $0.818 \pm 0.005$
& $0.858 \pm 0.009$
& $0.754 \pm 0.017$
& $0.697 \pm 0.009$ \\
\cmidrule(l){2-6}
& MIPROv2
& $0.797 \pm 0.006$
& $0.861 \pm 0.010$
& $0.739 \pm 0.019$
& $0.697 \pm 0.010$ \\
& GEPA
& $0.832 \pm 0.006$
& $0.878 \pm 0.011$
& $0.741 \pm 0.018$
& $0.704 \pm 0.010$ \\
& ACE
& $0.805 \pm 0.007$
& $0.889 \pm 0.011$
& $0.771 \pm 0.020$
& $0.718 \pm 0.011$ \\
& G\"{o}del Agent
& $0.835 \pm 0.024$
& $0.864 \pm 0.028$
& $0.780 \pm 0.043$
& $0.732 \pm 0.038$ \\
\cmidrule(l){2-6}
& +CE
& $0.835 \pm 0.019$
& $0.904 \pm 0.025$
& $0.752 \pm 0.036$
& $0.725 \pm 0.029$ \\
& +PE
& $0.832 \pm 0.007$
& $0.872 \pm 0.011$
& $0.789 \pm 0.018$
& $0.718 \pm 0.011$ \\
& \textbf{+PACE}
& $\mathbf{0.889 \pm 0.007}$
& $\mathbf{0.909 \pm 0.012}$
& $\mathbf{0.803 \pm 0.021}$
& $\mathbf{0.761 \pm 0.012}$ \\
\midrule
\multirow{8}{*}{\rotatebox{90}{Ministral-14B}}
& Vanilla
& $0.800 \pm 0.006$
& $0.770 \pm 0.008$
& $0.596 \pm 0.016$
& $0.781 \pm 0.009$ \\
\cmidrule(l){2-6}
& MIPROv2
& $0.833 \pm 0.007$
& $0.783 \pm 0.009$
& $0.583 \pm 0.018$
& $0.801 \pm 0.010$ \\
& GEPA
& $0.825 \pm 0.006$
& $0.775 \pm 0.010$
& $0.625 \pm 0.019$
& $0.816 \pm 0.011$ \\
& ACE
& $0.818 \pm 0.007$
& $0.785 \pm 0.011$
& $0.605 \pm 0.020$
& $0.788 \pm 0.012$ \\
& G\"{o}del Agent
& $0.797 \pm 0.027$
& $0.825 \pm 0.031$
& $0.659 \pm 0.045$
& $0.810 \pm 0.041$ \\
\cmidrule(l){2-6}
& +CE
& $0.836 \pm 0.022$
& $0.880 \pm 0.026$
& $0.654 \pm 0.035$
& $0.828 \pm 0.028$ \\
& +PE
& $0.832 \pm 0.007$
& $0.814 \pm 0.010$
& $0.709 \pm 0.022$
& $0.816 \pm 0.011$ \\
& \textbf{+PACE}
& $\mathbf{0.854 \pm 0.007}$
& $\mathbf{0.908 \pm 0.011}$
& $\mathbf{0.774 \pm 0.020}$
& $\mathbf{0.837 \pm 0.012}$ \\
\bottomrule
\end{tabular}}
\vspace{-3mm}
\end{table}

\subsection{Complete Experiment Results for Table~\ref{tab:main_results_std}}
We report the standard deviation of results in Table \ref{tab:main_results} in Table \ref{tab:main_results_std}. Three patterns stand out in Table~\ref{tab:main_results_std}. \emph{First}, PACE does not inflate variance despite its two-stage search: its standard deviation averages $0.011$ across the 12 backbone-benchmark cells, on par with +CE ($0.012$) and marginally above +PE ($0.009$).  The validation gate absorbs the instability of structural proposals rather than propagating it. \emph{Second}, structural methods (+CE, G\"{o}del Agent, +PACE) are consistently noisier than prompt-only baselines (+PE, GEPA, ACE, MIPROv2) on benchmarks where control-logic edits pay off (e.g.\ MGSM and HotpotQA on Qwen3.5-9B), reflecting run-to-run variability in \emph{which} strategy the SLM discovers, even when the expected accuracy is reliably higher. \emph{Third}, the main-text improvements survive these standard deviations.  On Qwen3.5-9B MMLU, +PACE beats +CE by $5.4$ points ($0.889{\pm}0.007$ vs.\ $0.835{\pm}0.008$), more than $3\sigma$ apart.  On Qwen3.5-9B IFEval, +PACE beats +PE by $4.3$ points ($0.761{\pm}0.012$ vs.\ $0.718{\pm}0.010$), roughly $2.7\sigma$.  The only narrow cluster is Qwen3-4B HotpotQA, where all methods sit within one $\sigma$ of Vanilla---consistent with our observation in Table \ref{tab:main_results} that the 4B model lacks the capacity for multi-hop reasoning regardless of agent design.

\subsection{Structural Edit Space}
\label{sec:edit-space}

Prompt evolution (+PE) and the outer loop of PACE share a common
interface, \texttt{action\_adjust\_logic}, but the categories of
change they are allowed to make on top of the current solver differ
qualitatively.  This subsection enumerates the concrete edit
categories that the outer-loop agent can emit, specifies whether each
is free-form Python or constrained to a template, and lists the
safety checks that gate every structural update before it is
committed.

\textbf{Interface.} Every structural edit is a single call of the form
\texttt{action\_adjust\_logic(module\_name, target\_name, new\_code,
target\_type, operation)} where
\texttt{operation}$\in$\{\texttt{modify}, \texttt{add},
\texttt{delete}\} and \texttt{target\_type}$\in$\{\texttt{function},
\texttt{class}\}.  By convention the principal target is the
solver function (\texttt{agent\_module.solver}); the same interface
can edit or introduce helper functions and classes within
\texttt{agent\_module}.  The prompt mutation interface
(\texttt{action\_optimize\_prompt\_on\_task}) is template-based and
operates only on the fields of the solver's
\texttt{prompt\_config}---role, requirements, temperature, and
response format---without touching control flow.

\textbf{Edit categories.}
Table~\ref{tab:edit-space} lists the edit categories the agent
actually uses across the four benchmarks, together with the
target, whether they are free-form or template-constrained, and
the safety checks that apply.  Templates reflect recurring patterns
in what the outer-loop agent discovers, but are not hard-coded:
the agent writes arbitrary Python that conforms to the solver
contract (\texttt{solver(agent, task)$\rightarrow$dict}).

\begin{table}[h]
\centering
\small
\caption{Allowed edit categories in the PACE outer loop.  All
edits are expressed as calls to \texttt{action\_adjust\_logic}.
\emph{Free-form} means the agent writes arbitrary Python satisfying
the solver contract; \emph{template-constrained} means the edit is
restricted to a fixed schema (prompt fields, temperature, or a
hyperparameter dict).}
\label{tab:edit-space}
\setlength{\tabcolsep}{4pt}
\renewcommand{\arraystretch}{1.15}
\begin{tabular}{@{}p{3.2cm} p{4.6cm} l p{3.4cm}@{}}
\toprule
\textbf{Category} & \textbf{Representative example} &
\textbf{Form} & \textbf{Safety constraints} \\
\midrule
Prompt field update
  & Change \texttt{role}, \texttt{requirements}, \texttt{temperature}, or \texttt{response\_format} in \texttt{prompt\_config}
  & Template
  & Schema-validated; no code execution \\
\addlinespace[2pt]
Sampling-policy change
  & Replace a single-call decode with $n$-way sampling at an altered temperature
  & Free-form
  & Syntax check; validation gate \\
\addlinespace[2pt]
Aggregation / selection
  & Add majority vote (MMLU), score-then-argmax (IFEval), or span-overlap selector (HotpotQA)
  & Free-form
  & Syntax check; validation gate \\
\addlinespace[2pt]
Early-exit gate
  & Return immediately on $\geq 2/3$ consensus or $s{=}1.0$ constraint pass
  & Free-form
  & Syntax check; validation gate \\
\addlinespace[2pt]
Self-verification step
  & Add a confirm-or-correct call at $T{=}0$ on the top candidate (MMLU)
  & Free-form
  & Syntax check; validation gate \\
\addlinespace[2pt]
Targeted repair / rewrite
  & Feed best candidate back as assistant turn and ask for a constraint-satisfying rewrite (IFEval)
  & Free-form
  & Syntax check; validation gate; acceptance gate ($s_{\text{repair}} \geq s_{\text{best}}$) \\
\addlinespace[2pt]
Output normalization
  & Strip articles, trailing \texttt{\%}, commas, or decorations from the answer string
  & Free-form
  & Syntax check; validation gate \\
\addlinespace[2pt]
Helper function / class
  & Introduce \texttt{self\_check\_constraints}, \texttt{extract\_answer}, or \texttt{normalize\_numeric}
  & Free-form
  & Must not modify \texttt{action\_call\_llm} or \texttt{action\_call\_json\_format\_llm} \\
\addlinespace[2pt]
Hyperparameter dict
  & Edit temperature bounds, sample count $n$, retry count, or early-exit threshold
  & Template
  & Schema-validated; validation gate \\
\addlinespace[2pt]
Deletion
  & Remove an unused helper or a branch that proved harmful
  & Free-form
  & Target must not be the solver, \texttt{action\_call\_llm}, or \texttt{action\_call\_json\_format\_llm} \\
\bottomrule
\end{tabular}
\end{table}

\textbf{Safety constraints.}
Every structural update is gated by a sequence of checks before it
is applied to \texttt{agent\_module}:
\begin{itemize}[nosep,leftmargin=*]
\item \textbf{Prompt-saturation gate.}  Structural updates are
      blocked unless $\Delta U_P < \varepsilon$ on the inner-loop
      credit state, i.e.\ the prompt optimizer has exhausted its
      gains.  This enforces the PE$\to$CE ordering described in the
      main body.
\item \textbf{Immutable targets.}  The LLM call primitives \texttt{Agent.action\_call\_llm} and \texttt{Agent.action\_call\_json\_format\_llm} cannot be edited or deleted.  Any attempt raises \texttt{ValueError} and the edit is discarded.
\item \textbf{Blacklisted primitives.}  The solver may not contain \texttt{time.sleep}, and no target may \texttt{import logging} or \texttt{from logging ...}.  These would break the evaluation harness' latency accounting and log capture.
\item \textbf{Syntax compilation.}  Every \texttt{modify} or \texttt{add} is parsed with \texttt{compile(...)}; a \texttt{SyntaxError} surfaces as a structured error message (line, column, offending text) and the edit is rejected.
\item \textbf{validation gate (solver only).}  Edits that modify the \texttt{solver} function are run through a cheap candidate comparison (\texttt{\_compare\_variant\_metrics}) against the current solver on $12$ deterministic validation samples.  If $\Delta_{\text{AB}} \leq \tau_{\text{AB}}$ (\texttt{agent.ab\_gate\_delta\_threshold}) the edit is rejected and the candidate code is not installed.
\item \textbf{Post-commit structural eval.}  When an edit passes the validation gate and is committed, the agent records the current eval accuracy as \texttt{structure\_eval\_baseline} and marks a pending full evaluation.  The next \texttt{action\_evaluate\_on\_task} computes $\Delta U_C$, which is fed back into the credit mechanism (Section~\ref{sec:failure-shift}) and can trigger rollback if the full-evaluation delta is negative.
\end{itemize}

\textbf{Implications.}
The edit space is deliberately asymmetric: prompt fields are small, template-constrained, and cheap to revert, while structural edits are free-form Python but guarded by a prompt-saturation gate, a validation gate, and a set of immutable primitives.  This is what lets a frozen 9B model act as its own architect---free-form enough to discover self-consistency on MMLU or the generate--score--repair loop on IFEval (Figures~\ref{fig:evolved-solver-mmlu}--\ref{fig:evolved-solver-ifeval}), while the gates prevent the model's occasional faulty edits from
regressing the system.

\subsection{Algorithm Walkthrough}
\label{sec:algorithm_walkthrough}
\begin{algorithm}[t]
\caption{PACE: Prompt And Control Logic Evolution}
\label{alg:pace}
\begin{algorithmic}[1]
\Require Frozen SLM $M_\theta$; initial prompt artifacts $P_0$; initial control logic $C_0$; 
training set $\mathcal{D}_{train}$; validation set $\mathcal{D}_{val}$; resource budget $B$; 
prompt saturation threshold $\epsilon$; validation gate acceptance margin $\delta$; 
maximum outer-loop steps $K$; maximum prompt-evolution steps $L$
\Ensure Evolved agent $A^* = (P^*, C^*)$

\State Initialize $P \leftarrow P_0$, $C \leftarrow C_0$
\State Evaluate initial agent $A=(P,C)$ on $\mathcal{D}_{val}$ to obtain $U_{best}$

\For{$k = 1$ to $K$}
    \Statex \Comment{\textbf{Inner loop: prompt evolution under fixed control logic}}
    \State $P_{\mathrm{old}} \leftarrow P$
    \For{$\ell = 1$ to $L$}
        \State Generate prompt candidates $\mathcal{P}_{cand}$ using handcrafted mutations, failure-guided reflection, and crossover
        \State Evaluate each $P' \in \mathcal{P}_{cand}$ with fixed $C$ on minibatches from $\mathcal{D}_{train}$
        \State Update the prompt Pareto front according to validation utility and resource cost
        \State Select the best prompt $P$ from the Pareto front using $\mathcal{D}_{val}$
        \If{$Cost(P,C) > B$}
            \State Reject $P$ and restore the best feasible prompt on the Pareto front
        \EndIf
    \EndFor

    \State Compute recent prompt improvement 
    $\Delta U_P \leftarrow U(P,C;\mathcal{D}_{val}) - U(P_{\mathrm{old}},C;\mathcal{D}_{val})$

    \If{$\Delta U_P \geq \epsilon$}
        \State \textbf{continue} \Comment{Prompt evolution is still useful; do not edit control logic}
    \EndIf

    \Statex \Comment{\textbf{Outer loop: constrained structural evolution}}
    \State Analyze failures of $(P,C)$ on $\mathcal{D}_{val}$
    \State Classify failures into structural categories, e.g., parsing errors, validation failures, retry failures, or inference-configuration issues
    \State Generate a constrained structural candidate $C'$ within the safe-to-edit search space

    \If{$Cost(P,C') > B$}
        \State Reject $C'$
        \State \textbf{continue}
    \EndIf

    \Statex \Comment{\textbf{Empirical validation validation}}
    \State Evaluate old agent $A_{old}=(P,C)$ and candidate agent $A_{new}=(P,C')$ on the same validation subset $\mathcal{V} \subseteq \mathcal{D}_{val}$
    \State Compute $\Delta U_C \leftarrow U(A_{new};\mathcal{V}) - U(A_{old};\mathcal{V})$

    \If{$\Delta U_C > \delta$}
        \State Accept the structural update: $C \leftarrow C'$
        \State Re-enter prompt evolution under the updated control logic
    \Else
        \State Reject $C'$ and keep $C$ unchanged
    \EndIf

    \State Update $U_{best} \leftarrow U(P,C;\mathcal{D}_{val})$ if improved
\EndFor

\State \Return $A^*=(P,C)$
\end{algorithmic}
\end{algorithm}

Algorithm 1 summarizes the full PACE procedure. Starting from an initial agent, PACE first performs prompt evolution while holding the control logic fixed. Prompt candidates are generated through mutation, reflection, and crossover, then selected using validation utility and resource cost. If the marginal validation improvement from prompt evolution remains above the saturation threshold $\epsilon$, PACE continues refining prompts. Once prompt gains fall below $\epsilon$, the framework activates constrained structural evolution. Structural candidates are proposed only within a predefined safe-to-edit search space and are accepted only if they improve over the current agent by at least the validation margin $\delta$ on the same held-out validation subset without violating the resource budget. After each accepted structural update, PACE returns to prompt evolution, allowing the prompt to adapt to the new control logic.


\subsection{Token Usage Analysis}
\label{sec:token-analysis}

A defining constraint of the SLM setting is that every token is generated locally on limited hardware.  Unlike API-based systems where cost is monetary, here cost is \emph{latency and throughput}: each additional LLM call occupies a GPU that could serve another request. We therefore analyse the token budget of both the evolution process (outer loop) and the evolved solvers (inference time) to characterise the practical overhead of self-evolution under resource constraints.%
\footnote{Token counts in this section are estimates derived from code-level analysis of prompt templates, solver call patterns, and observed output lengths.  The current implementation tracks cost via \texttt{len(response\_text)} (character count) rather than API-level \texttt{usage.prompt\_tokens} / \texttt{completion\_tokens}.  Accuracy figures are from real evaluation runs on the respective test splits (800 samples for MMLU, 141 for IFEval).}

\begin{table}[t]
\centering
\small
\caption{Estimated per-query inference token breakdown for baseline and evolved solvers.  \emph{Input} counts prompt tokens (system + user + context); \emph{Output} counts generated tokens.  Multipliers are relative to the single-call baseline. Token counts are derived from code-level analysis of prompt templates and observed output lengths (marked with ${\sim}$); accuracy figures are from real evaluation runs on the respective test splits.}
\label{tab:token-breakdown}
\setlength{\tabcolsep}{5pt}
\begin{tabular}{@{}l l r r r c@{}}
\toprule
\textbf{Benchmark} & \textbf{Solver variant}
  & \textbf{Input} & \textbf{Output} & \textbf{Total}
  & \textbf{Multiplier} \\
\midrule
\multirow{3}{*}{MMLU}
  & Baseline (single call)
    & ${\sim}$350 & ${\sim}$120 & ${\sim}$470 & 1.0$\times$ \\
  & +PE (optimized prompt)
    & ${\sim}$420 & ${\sim}$150 & ${\sim}$570 & 1.2$\times$ \\
  & +PACE (voting + verify)
    & ${\sim}$1\,260 & ${\sim}$450 & ${\sim}$1710 & 3.6$\times$ \\
\midrule
\multirow{3}{*}{IFEval}
  & Baseline (single call)
    & ${\sim}$280 & ${\sim}$350 & ${\sim}$630 & 1.0$\times$ \\
  & +PE (optimized prompt)
    & ${\sim}$340 & ${\sim}$380 & ${\sim}$720 & 1.1$\times$ \\
  & +PACE (score + repair)
    & ${\sim}$1700 & ${\sim}$1500 & ${\sim}$3200 & 5.1$\times$ \\
\midrule
\multirow{3}{*}{MGSM}
& Baseline (single call)& ${\sim}$130 & ${\sim}$160 & ${\sim}$290 & 1.0$\times$ \\
& +PE (optimized prompt)& ${\sim}$275 & ${\sim}$200 & ${\sim}$475 & 1.6$\times$ \\
& +PACE (translate + vote + verify)& ${\sim}$420 & ${\sim}$1\,100 & ${\sim}$1\,520 & 5.2$\times$ \\
\midrule
\multirow{3}{*}{HotpotQA}
& Baseline (single call)& ${\sim}$1\,565 & ${\sim}$180 & ${\sim}$1\,745 & 1.0$\times$ \\
& +PE (optimized prompt)& ${\sim}$1\,780 & ${\sim}$200 & ${\sim}$1\,980 & 1.1$\times$ \\
& +PACE (answer refinement)& ${\sim}$2\,390 & ${\sim}$320 & ${\sim}$2\,710 & 1.6$\times$ \\
\bottomrule
\end{tabular}
\vspace{-4mm}
\end{table}

\textbf{Inference-time token cost.}
Table~\ref{tab:token-breakdown} decomposes the per-query token budget
across solver variants, and two patterns stand out:

\begin{itemize}[nosep,leftmargin=*]
  \item \textit{Prompt evolution is nearly free.}  The optimized prompt adds only 50--70 tokens of system-prompt overhead (longer role descriptions and requirement suffixes), yielding a 1.1--1.2$\times$ multiplier with no additional LLM calls.  This makes PE the highest-ROI intervention: on MMLU it accounts for the majority of the accuracy gain at negligible cost.

 \item \textit{Structural evolution trades tokens for accuracy.} The evolved MMLU solver issues up to $3{+}1$ calls (3-way sample $+$ conditional verification), while the IFEval solver issues up to $3{+}3{+}1$ calls (3-way sample + 3 self-checks + conditional repair).  The MMLU multiplier is kept moderate (${\sim}$3.6$\times$) because the early-exit on consensus fires ${\sim}$80\% of the time, avoiding the verification call.  IFEval is more expensive (${\sim}$5.1$\times$) because the self-check scoring requires a separate LLM call per candidate, and the repair pass generates a full-length response.  However, the accuracy gain on IFEval is substantially larger (+33.1~pp from +PE to +PACE), reflecting the high value of structural intervention for constrained-generation tasks.  MGSM exhibits the highest per-query multiplier (${\sim}$5.2$\times$) due to its translate-then-solve pipeline plus $N{=}5$ self-consistency voting, but this cost is justified: majority voting over diverse reasoning paths catches stochastic arithmetic errors that no single prompt can eliminate, yielding +14.5~pp on Ministral-14B.  In contrast, HotpotQA's evolved solver adds only a lightweight second extraction round with question-type-aware routing (${\sim}$1.6$\times$), because the dominant failure mode---answer over-specification---is largely addressable through prompt refinement alone; the structural contribution is a yes/no detector that forces exact ``yes''/``no'' output for comparison questions, eliminating a failure mode worth ${\sim}$3~F1 points that prompt wording alone could not resolve.

\end{itemize}

\noindent Crucially, these multipliers are \emph{discovered} by the agent, not prescribed.  The Pareto-front optimization in the inner loop and the validation gate in the outer loop jointly penalize candidates that increase cost without proportional accuracy gain, steering the agent toward architectures with built-in cost control (early exits, conditional refinement).

\begin{table}[t]
\centering
\small
\caption{Estimated total token budget for the evolution process (outer loop), broken down by phase.  All figures are for Qwen3.5-9B with $K_{\max}{=}20$ outer steps,  extrapolated from code-level analysis of
a single evolution run.}\label{tab:evolution-budget}
\setlength{\tabcolsep}{5pt}
\begin{tabular}{@{}l r r r r r l@{}}
\toprule
\textbf{Phase} & \textbf{MMLU} & \textbf{IFEval}  & \textbf{MGSM} & \textbf{HpQA} & \textbf{Description} \\
\midrule
Prompt optimisation    & ${\sim}$1.2M & ${\sim}$1.8M  & ${\sim}$1.2M & ${\sim}$2.4M & Inner-loop candidates $\times$ train/valid evals \\
Evaluation calls       & ${\sim}$0.4M & ${\sim}$0.6M  & ${\sim}$0.5M & ${\sim}$0.5M & Mini-batch (50 samples) + test-set evals \\
Outer-loop reasoning   & ${\sim}$0.3M & ${\sim}$0.3M  & ${\sim}$0.4M & ${\sim}$0.6M &  Conversation, tool calls, failure analysis \\
Structural exploration & ${\sim}$0.2M & ${\sim}$0.3M  & ${\sim}$0.4M & ${\sim}$0.4M & validation comparisons, solver executions \\
\midrule
\textbf{Total evolution} & ${\sim}$\textbf{2.1M} & ${\sim}$\textbf{3.0M} & ${\sim}$\textbf{2.5M} & ${\sim}$\textbf{3.9M}  & One-time, amortised over future queries \\
\bottomrule
\end{tabular}
\end{table}

\textbf{Evolution-time token budget.}
Table~\ref{tab:evolution-budget} reports the estimated total tokens consumed during a single evolution run.  The dominant cost is prompt optimization (${\sim}$60\% of the budget), which evaluates multiple candidate configurations on train and validation splits across several iterations.  The outer-loop reasoning (agent conversation history, tool-call arguments, failure analysis summaries) accounts for only ${\sim}$15\%, demonstrating that the framework's token-trimming
mechanism (\texttt{\_trim\_messages\_to\_budget}) effectively controls context growth over long evolution trajectories.

The total evolution budget of 2--3M tokens is a \emph{one-time} cost that produces a permanently improved solver.  For context, this is equivalent to ${\sim}$3\,000--5\,000 inference queries under the evolved solver---a break-even point reached quickly in any deployment scenario.

\begin{table}[t!]
\centering
\small
\caption{Accuracy--cost Pareto trajectories for MMLU and IFEval using Qwen3.5-9B. Each row represents a solver variant: baseline, prompt-evolved (+PE), and fully evolved (+PACE). $\Delta$Acc reports the incremental gain over the previous row within each benchmark. Efficiency is the incremental accuracy gain per 1,000 additional tokens. Accuracy values are taken from Table~\ref{tab:main_results}; token counts are estimates.}
\label{tab:accuracy_cost}
\resizebox{0.78\textwidth}{!}{
\begin{tabular}{llcccc}
\toprule
\textbf{Benchmark} & \textbf{Variant}
& \textbf{Tokens/query}
& \textbf{Accuracy}
& \textbf{$\Delta$Acc (pp)}
& \textbf{Eff. (pp/1k tok)} \\
\midrule

\multirow{3}{*}{MMLU}
& Baseline
& 470
& 0.818
& --
& -- \\
& +PE
& 570
& 0.832
& +1.4
& 14.0 \\
& +PACE
& 1710
& 0.889
& +5.7
& 5.0 \\
\midrule

\multirow{3}{*}{IFEval}
& Baseline
& 630
& 0.697
& --
& -- \\
& +PE
& 720
& 0.718
& +2.1
& 23.3 \\
& +PACE
& 3200
& 0.761
& +4.3
& 1.7 \\

\midrule
\multirow{3}{*}{MGSM}
& Baseline& 290& 0.858& --& -- \\
& +PE& 475& 0.872& +1.3& 7.0 \\
& +PACE& 1\,520& 0.909& +5.1& 4.1 \\
\midrule
\multirow{3}{*}{HotpotQA}
& Baseline& 1\,745& 0.754& --& -- \\
& +PE& 1\,980& 0.789& +3.5& 14.9 \\
& +PACE & 2\,710& 0.803& +4.9& 5.0 \\
\bottomrule
\end{tabular}}
\vspace{-3mm}
\end{table}

\textbf{Why token efficiency matters for SLMs.}
The token analysis reveals a cost structure qualitatively different
from the frontier-model setting. First, \textit{output tokens dominate}: SLMs produce longer reasoning chains than frontier models to reach comparable accuracy, making output-token efficiency a first-order concern that the Pareto-front
optimization directly addresses. Second, \textit{batched sampling amortizes latency}: both evolved solvers use $n{>}1$ in a single API call; on vLLM, batched sampling with $n{=}3$ is only ${\sim}1.3$--$1.5{\times}$ slower than $n{=}1$
thanks to KV-cache sharing, so the cost sensitivity of the validation gate implicitly favors batched over sequential designs. Third, the \textit{evolution cost is a fixed upfront investment} (2--3M tokens); once the solver is evolved it runs at the per-query cost in Table~\ref{tab:token-breakdown} indefinitely, an amortization
property especially attractive for SLM deployments serving many queries on dedicated hardware. Finally, \textit{credit assignment controls cost growth}: the
$\varepsilon$-gated PE$\to$CE transition (Section~\ref{sec:ablation}) exhausts cheap prompt-level gains before incurring multi-call structural changes, avoiding the premature cost inflation visible in the $\varepsilon{=}1.0$ row of
Table~\ref{tab:ablation}.

\textbf{Comparison with frontier-model baselines.}
The evolved Qwen3.5-9B solver uses ${\sim}$1\,710 tokens but runs on a single A100 GPU at zero marginal API cost. At a serving throughput of ${\sim}$2\,000 tokens/s, the evolved solver adds ${\sim}$0.6\,s of latency per query, which is acceptable for batch evaluation and many interactive use cases.  This positions SLM self-evolution as a practical alternative to frontier-model API access: a modest one-time compute investment yields a permanently improved local model that approaches frontier accuracy on structured benchmarks without ongoing API costs.

\subsection{Cross-Benchmark Solver Analysis: MMLU vs.\ IFEval}
\label{sec:solver-comparison}

The structural evolution trajectories on MMLU and IFEval reveal that the outer-loop agent independently discovers qualitatively different inference strategies tailored to each task's evaluation semantics, without any human-specified architectural prior.  We analyse the two evolved solvers side by side to highlight how a frozen SLM, acting as its own architect, adapts control logic to the structure of the problem.

\begin{table}[t]
\centering
\small
\caption{Structural comparison of evolved solvers on MMLU and IFEval.
  Both are discovered autonomously by the same outer-loop agent
  (Qwen3.5-9B) via \texttt{action\_adjust\_logic}.}
\label{tab:solver-comparison}
\setlength{\tabcolsep}{4pt}
\begin{tabular}{@{}l p{5.0cm} p{5.0cm}@{}}
\toprule
\textbf{Dimension}
  & \textbf{MMLU (classification)}
  & \textbf{IFEval (constrained generation)} \\
\midrule
Answer space
  & Closed set $\{A,B,C,D\}$
  & Open-ended text subject to $k$ verifiable constraints \\
\addlinespace[2pt]
Phase 1: Diversity
  & $n{=}3$ samples at $T{\geq}0.5$; extract discrete labels
  & $n{=}3$ samples at $T{\geq}0.3$; retain full response text \\
\addlinespace[2pt]
Selection criterion
  & Majority vote (${\geq}2/3$ consensus)
  & Constraint-satisfaction score $s \in [0,1]$; pick $\arg\max_i s_i$ \\
\addlinespace[2pt]
Early exit
  & Consensus $\Rightarrow$ return immediately
  & $s{=}1.0$ (all constraints pass) $\Rightarrow$ return immediately \\
\addlinespace[2pt]
Phase 2: Refinement
  & Verification tiebreak: present top answer + reasoning, ask
    model to confirm or correct at $T{=}0$
  & Targeted repair: feed best candidate back as assistant turn,
    ask model to rewrite satisfying all constraints at $T{=}0$ \\
\addlinespace[2pt]
Acceptance gate
  & Accept verified answer if valid label
  & Accept repair only if $s_{\text{repair}} \geq s_{\text{best}}$ \\
\addlinespace[2pt]
Output format
  & JSON with \texttt{reasoning}, \texttt{answer} keys
  & Raw text (\texttt{response\_format="text"}) to avoid
    format-constraint conflicts \\
\addlinespace[2pt]
Avg.\ LLM calls
  & ${\sim}1.3\times$ (3-way batch + 15--20\% verification)
  & ${\sim}2.0\times$ (3-way batch + scoring + conditional repair) \\
\bottomrule
\end{tabular}
\end{table}

\textbf{Shared meta-strategy.}
Despite targeting fundamentally different evaluation protocols, both evolved solvers converge on the same three-phase skeleton: \emph{diverse generation} $\rightarrow$ \emph{selection} $\rightarrow$ \emph{conditional refinement}.  This is not prescribed by the framework; the agent arrives at it independently on each benchmark. The convergence suggests that \emph{generate--evaluate--refine} is a natural attractor in the space of inference-time strategies discoverable by SLMs, analogous to how self-consistency and chain-of-thought emerge as effective patterns in the prompting literature~\citep{wang2023selfconsistency,wei2022chain}.

\textbf{Task-adaptive specialization.}
Within the shared skeleton, the agent makes task-specific design choices that a human engineer would recognize as sensible:

\begin{itemize}[nosep,leftmargin=*]
  \item \textbf{Selection mechanism.}  MMLU admits a trivial
    aggregation (majority vote over four labels), whereas IFEval
    requires evaluating each candidate against a heterogeneous
    constraint set.  The agent replaces voting with an LLM-as-judge self-check that enumerates constraints and scores compliance--- effectively inventing a lightweight verifier without access to the ground-truth checker.
  \item \textbf{Refinement strategy.}  For MMLU, a wrong answer is best addressed by re-examining the reasoning chain (verification). For IFEval, a constraint violation is best addressed by rewriting the offending passage (repair).  The agent discovers this distinction: it uses a \emph{confirm-or-correct} prompt for MMLU and a \emph{rewrite-to-satisfy} prompt for IFEval.
  \item \textbf{Output format.}  The MMLU solver requests structured JSON to facilitate answer extraction, while the IFEval solver explicitly uses \texttt{response\_format="text"} to avoid introducing formatting artefacts that would violate content constraints.  This choice reflects awareness of the interaction between output format and evaluation criteria.
  \item \textbf{Temperature regime.}  The MMLU solver uses
    $T{\geq}0.5$ for diversity (reasoning paths are more varied at higher temperature), while the IFEval solver uses $T{\geq}0.3$ (lower diversity suffices because constraint violations are structural, not stochastic).
\end{itemize}

\begin{figure}[t]
\centering
\begin{tcolorbox}[
  colback=gray!3,
  colframe=black!60,
  fontupper=\ttfamily\scriptsize,
  left=4pt, right=4pt, top=3pt, bottom=3pt,
  title={\small Evolved MMLU solver (self-consistency + verification)},
  boxrule=0.4pt,
]
\begin{alltt}
\textbf{def} solver(agent, task):
    cfg = normalize\_prompt\_config(agent.prompt\_config)
    msg = [\{"role": "user", "content": task\}]

    \textrm{\textit{\color{gray}# Phase 1: 3-way self-consistency (temp >= 0.5)}}
    resps = call\_llm(msg, temp=max(cfg["T"], 0.5), n=3)
    votes, best = \{\}, \{\}
    \textbf{for} r \textbf{in} resps:
        a = extract\_answer(r)          \textrm{\textit{\color{gray}# coerce to A/B/C/D}}
        \textbf{if} a \textbf{in} \{"A","B","C","D"\}:
            votes[a] = votes.get(a, 0) + 1
            best.setdefault(a, r)

    top = max(votes, key=votes.get)
    \textbf{if} votes[top] >= 2:              \textrm{\textit{\color{gray}# consensus}}
        \textbf{return} best[top]

    \textrm{\textit{\color{gray}# Phase 2: verification tiebreak (temp = 0)}}
    v = call\_llm(
        f"Previous answer: \{top\}\\nReasoning: "
        f"\{best[top]['reasoning'][:600]\}\\n"
        "Confirm or correct.", temp=0.0, n=1)
    \textbf{return} extract\_answer(v[0]) or best[top]
\end{alltt}
\end{tcolorbox}
\caption{Simplified evolved solver for MMLU.  \texttt{call\_llm} wraps
  \texttt{action\_call\_json\_format\_llm}; \texttt{extract\_answer}
  wraps the coercion and regex extraction pipeline.}
\label{fig:evolved-solver-mmlu}
\end{figure}

\begin{figure}[t]
\centering
\begin{tcolorbox}[
  colback=gray!3,
  colframe=black!60,
  fontupper=\ttfamily\scriptsize,
  left=4pt, right=4pt, top=3pt, bottom=3pt,
  title={\small Evolved IFEval solver (constraint scoring + targeted repair)},
  boxrule=0.4pt,
]
\begin{alltt}
\textbf{def} solver(agent, task):
    cfg = normalize\_prompt\_config(agent.prompt\_config)
    sys = build\_system\_prompt(cfg)
    msg = [\{"role":"system","content":sys\},
           \{"role":"user","content":task\}]

    \textrm{\textit{\color{gray}# Phase 1: generate 3 candidates (temp >= 0.3)}}
    resps = call\_llm(msg, temp=max(cfg["T"], 0.3),
                      n=3, fmt="text")
    candidates = [extract\_text(r) \textbf{for} r \textbf{in} resps]

    \textrm{\textit{\color{gray}# Phase 2: pick best via constraint self-check}}
    best, best\_score = None, -1.0
    \textbf{for} c \textbf{in} candidates:
        s = self\_check\_constraints(agent, task, c)
        \textbf{if} s > best\_score:
            best, best\_score = c, s
    \textbf{if} best\_score >= 1.0:          \textrm{\textit{\color{gray}# all constraints pass}}
        \textbf{return} \{"response": best\}

    \textrm{\textit{\color{gray}# Phase 3: targeted repair (temp = 0)}}
    repair\_msg = msg + [
        \{"role":"assistant","content":best\},
        \{"role":"user","content":
         "The response above violates some constraints."
         " Rewrite so EVERY constraint is satisfied."
         " Output ONLY the corrected response."\}]
    fixed = call\_llm(repair\_msg, temp=0.0, n=1, fmt="text")

    \textrm{\textit{\color{gray}# Accept repair only if it improves}}
    fix\_s = self\_check\_constraints(
        agent, task, extract\_text(fixed[0]))
    \textbf{if} fix\_s >= best\_score:
        \textbf{return} \{"response": extract\_text(fixed[0])\}
    \textbf{return} \{"response": best\}
\end{alltt}
\end{tcolorbox}
\caption{Simplified evolved solver for IFEval.
  \texttt{self\_check\_constraints} prompts the model to enumerate
  constraints and score compliance; \texttt{call\_llm} wraps
  \texttt{action\_call\_llm} with \texttt{response\_format="text"}.
  Compare with the MMLU solver in Figure~\ref{fig:evolved-solver-mmlu}.}
\label{fig:evolved-solver-ifeval}
\end{figure}

\paragraph{Implications for SLM-driven self-evolution.}
The cross-benchmark comparison surfaces several properties that distinguish SLM self-evolution from conventional prompt engineering or human-designed inference pipelines:

\begin{enumerate}[nosep,leftmargin=*]
  \item \textbf{Emergent architectural transfer.}  The
    generate--select--refine skeleton emerges independently on both tasks, suggesting that SLMs possess implicit knowledge of effective inference-time compute patterns.  This is notable because
    the 9B model was never explicitly trained on meta-reasoning about solver design; the pattern is \emph{discovered} through the interplay of failure analysis, credit assignment, and code generation.

  \item \textbf{Constraint-aware adaptation without oracle access.}
    On IFEval, the agent invents a self-check scoring mechanism that approximates the ground-truth verifier
    (\texttt{verify\_all\_instructions}) without ever observing it. The self-check is imperfect---it relies on the same frozen model that generated the response---but it provides a sufficient signal to rank candidates and gate repairs.  This demonstrates that SLMs can construct \emph{task-specific evaluation proxies} as part of structural evolution, partially compensating for the absence of external reward models.

  \item \textbf{Cost-aware design under implicit budgets.}
    Both solvers include early-exit conditions (consensus for MMLU, full-pass for IFEval) that avoid unnecessary refinement calls. The agent is not given an explicit cost objective, yet it
    discovers short-circuit logic that reduces average inference cost. This suggests that the validation gate mechanism, which rejects candidates that increase cost without proportional accuracy gain, implicitly teaches the agent to prefer efficient architectures.

  \item \textbf{Robustness through conservative acceptance.}
    The IFEval solver accepts a repair only if it scores at least as well as the best candidate, preventing the repair pass from introducing new violations.  Similarly, the MMLU solver falls back to the plurality answer when verification produces an invalid label.  Both patterns reflect a \emph{do-no-harm} principle that the agent learns from the empirical validation loop: edits that cannot be shown to help are rejected.
\end{enumerate}

\paragraph{Unique strengths of SLM-powered evolution.}
Taken together, these observations highlight capabilities that are distinctive to the self-evolving SLM paradigm:

\begin{itemize}[nosep,leftmargin=*]
  \item \textbf{Zero-shot solver design.}  The agent synthesises multi-phase inference strategies from scratch, without few-shot examples of solver code or access to a library of known techniques.  A 9B model autonomously rediscovers self-consistency~\cite{wang2023selfconsistency} for MMLU and  invents a generate--score--repair loop for IFEval---strategies that required separate research efforts when designed by humans.

  \item \textbf{Benchmark-specific structural creativity.}  Rather than applying a single universal strategy, the agent tailors control flow, output format, temperature, and refinement logic to the evaluation protocol of each benchmark.  This level of task-specific adaptation is difficult to achieve with prompt-only methods or fixed inference pipelines.

  \item \textbf{Self-improving without stronger teachers.}  The entire evolution loop---prompt mutation, failure analysis, code generation, and empirical validation---is executed by the \emph{same} frozen SLM that serves as the inference engine.  No larger teacher model, human feedback, or gradient updates are required.  The improvement comes purely from better \emph{use} of the model's existing capabilities through evolved control logic.

  \item \textbf{Graceful degradation via empirical gating.}  The credit-assignment and validation gate mechanisms ensure that structural exploration is both \emph{triggered} and \emph{validated} empirically.  When the SLM proposes a flawed edit---which happens frequently at 9B scale---the gate rejects it without degrading the current solver.  This makes the evolution process monotonically non-regressing in expectation, a property that is critical for deploying autonomous agents in practice.
\end{itemize}

Table~\ref{tab:solver-comparison} and
Figures~\ref{fig:evolved-solver-mmlu}--\ref{fig:evolved-solver-ifeval} together illustrate that the PACE framework enables a single frozen SLM to act as both the \emph{subject} and the \emph{architect} of its own inference pipeline, discovering task-appropriate strategies that would traditionally require separate human engineering effort for each benchmark.


\subsection{$\tau$-bench Evolution Analysis}\label{app:tau-analysis}

We provide a detailed analysis of the PACE evolution trajectory on $\tau$-bench Retail using Qwen3.5-9B, tracing both the failure taxonomy shifts and the accuracy progression across evolution phases.

\subsubsection{Failure Taxonomy}

Table~\ref{tab:tau-failure-taxonomy} reports the failure category breakdown at each evolution checkpoint.
The $\tau$-bench evaluation framework classifies failures into four categories:
\emph{missing\_actions} (the agent omitted a required tool call),
\emph{extra\_actions} (the agent invoked an unnecessary or incorrect tool),
\emph{tool\_errors} (the agent called a tool with invalid arguments, e.g., querying a non-existent user),
and \emph{early\_stop} (the user simulator terminated the conversation prematurely due to agent confusion or unresponsiveness).

\begin{table}[h]
\centering
\small
\caption{Failure taxonomy across evolution phases for Qwen3.5-9B on $\tau$-bench Retail (115 tasks, single-trial mid-loop evaluation). Categories are non-exclusive: a single failed task may exhibit multiple failure types (e.g., an extra tool call that also triggers a tool error and leads to early user termination).}

\label{tab:tau-failure-taxonomy}
\setlength{\tabcolsep}{5pt}
\begin{tabular}{@{}lcccccc@{}}
\toprule
\textbf{Phase} & \textbf{Pass} & \textbf{Fail} & \textbf{Missing} & \textbf{Extra} & \textbf{Tool Err} & \textbf{Early Stop} \\
\midrule
Baseline (vanilla)       & 90 & 25 & 9  & 12 & 9 & 18 \\
+ PE       & 93 & 22 & 9  & 10 & 11 & 21 \\
+ PACE w/ structural edit (best) & 96 & 19 & 7  & 11 & 4  & 16 \\
\bottomrule
\end{tabular}
\end{table}

Three patterns emerge from the taxonomy progression:

\begin{enumerate}[nosep,leftmargin=*]
\item \emph{Tool errors are halved by structural evolution.}
Tool errors increase slightly under PE (9$\to$11) but drop sharply to 4 after the structural edit, representing the largest relative reduction among all categories.
This confirms that structural evolution addresses failure modes that prompt refinement cannot: the evolved control logic adds pre-call validation (e.g., verifying order status before executing irreversible actions) and enforces the single-call-per-order constraint that the SLM frequently violates under prompt-only guidance.

\item \emph{Early stop remains the dominant failure mode throughout.}
Early termination accounts for 70--80\% of failures at every checkpoint.
These failures arise when the stochastic user simulator loses patience with the agent's responses --- a signal that is difficult to address through either prompt or structural changes alone, as it depends on the model's conversational fluency and the simulator's tolerance threshold.

\begin{figure}[t]
\centering
\includegraphics[width=0.85\linewidth]{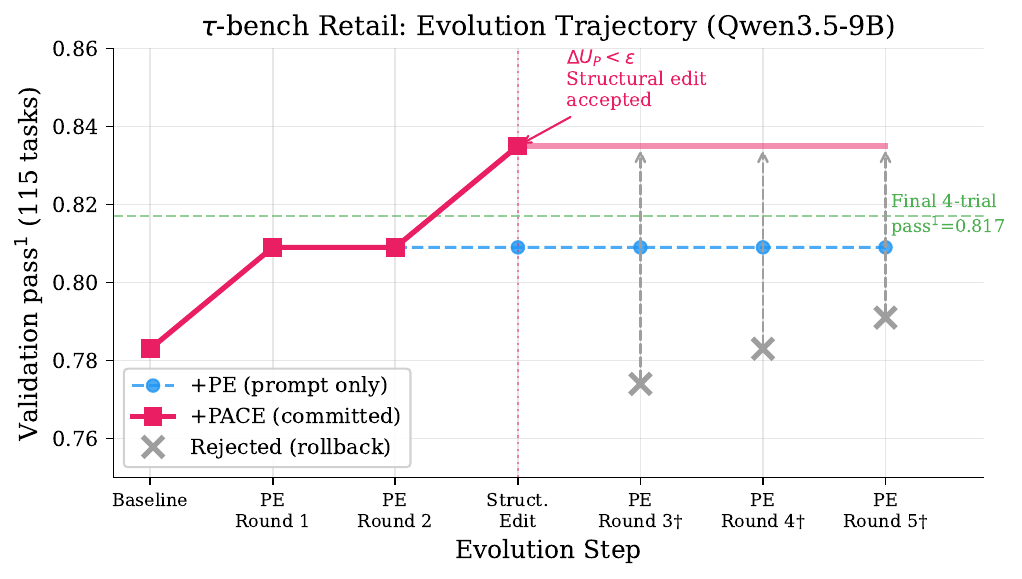}
\caption{Evolution trajectory for Qwen3.5-9B on $\tau$-bench Retail (115 tasks, single-trial validation during evolution). The +PE baseline (blue, dashed) saturates at $0.809$ after two rounds. PACE (red, solid) matches PE during the prompt phase, then achieves a discrete jump to $0.835$ when the structural edit is accepted at step~3. Subsequent PE rounds (gray crosses) regress and are rejected by the validation gate, preserving the post-structural peak. The green dashed line marks the final 4-trial $\text{pass}^1{=}0.817$.}
\label{fig:tau-trajectory}
\end{figure}

\item \emph{Missing actions decrease only after structural intervention.}
Prompt evolution leaves missing\_actions unchanged (9$\to$9), while the structural edit reduces it to 7.
The evolved solver's requirement to ``collect ALL items into a single list before making the call'' prevents the common failure pattern where the agent processes items sequentially and exhausts the one-call-per-order limit before completing all user requests.
\end{enumerate}

\subsubsection{Evolution Trajectory}

Figure~\ref{fig:tau-trajectory} visualizes the full evolution trajectory.

The trajectory reveals several characteristics specific to multi-turn agent evolution:

\begin{enumerate}[nosep,leftmargin=*]
\item \emph{A single structural edit provides the largest discrete jump.}
The accepted edit (step~3) improves validation accuracy from $0.809$ to $0.835$ (+2.6~pp), the largest single-step gain in the trajectory.
The edit adds tool-call pre-validation and enforces the single-call-per-order constraint, directly addressing the \texttt{tool\_errors} and \texttt{missing\_actions} categories identified in the failure taxonomy.

\item \emph{The validation gate prevents post-peak regression.}
Steps 4--6 show that subsequent PE rounds after the structural edit consistently produce worse results ($0.774$, $0.783$, $0.791$), all rejected by the rollback mechanism.
This demonstrates the monotonic non-regression property of PACE: once a good structural configuration is found, the gate prevents destabilizing changes.

\item \emph{Multi-trial evaluation is lower than single-trial mid-loop.}
The final 4-trial evaluation ($\text{pass}^1{=}0.817$) is lower than the best validation score ($0.835$).
This gap reflects the inherent variance of multi-turn dialogue with a stochastic user simulator: a solver that passes 96/115 tasks in one trial may fail different subsets in subsequent trials.
The $\text{pass}^4$ improvement over vanilla ($0.661$ vs.\ $0.626$) confirms that PACE's structural edits improve robustness across trials, not just peak single-run performance.
\end{enumerate}



\end{document}